\documentclass{article}
\usepackage{ijcai24}

\usepackage{times}
\usepackage{soul}
\usepackage{url}
\usepackage[hidelinks]{hyperref}
\usepackage[utf8]{inputenc}
\usepackage[small]{caption}
\usepackage{graphicx}
\usepackage{amsmath}
\usepackage{amsthm}
\usepackage{booktabs}
\usepackage{algorithm}
\usepackage{algorithmic}
\usepackage[switch]{lineno}
\usepackage{amsfonts}
\usepackage{array}
\usepackage{multirow}
\usepackage{makecell}
\title{ClipSAM: CLIP and SAM Collaboration for Zero-Shot Anomaly Segmentation}

\author{
Shengze Li$^1$\and
Jianjian Cao$^1$\and
Peng Ye$^{1}$\and
Yuhan Ding$^1$\and
Chongjun Tu$^1$\And
Tao Chen$^1$\footnote{Corresponding author}\\
\affiliations
$^1$School of Information Science and Technology, Fudan University\\
}
\begin{document}

\maketitle

\begin{abstract}

%chentao
Recently, foundational models such as CLIP and SAM have shown promising performance for the task of Zero-Shot Anomaly Segmentation (ZSAS). 
However, either CLIP-based or SAM-based ZSAS methods still suffer from non-negligible key drawbacks: 1) CLIP primarily focuses on global feature alignment across different inputs, leading to imprecise segmentation of local anomalous parts; 2) SAM tends to generate numerous redundant masks without proper prompt constraints, resulting in complex post-processing requirements. 
% 1) CLIP focuses on global rather than local feature alignment of different inputs, resulting in imprecise segmentation of anomalous local parts; 2) SAM tends to generate numerous redundant masks without prompted constraints, imposing great post-processing complexities. 
In this work, we innovatively propose a CLIP and SAM collaboration framework called ClipSAM for ZSAS. The insight behind ClipSAM is to employ CLIP's semantic understanding capability for anomaly localization and rough segmentation, which is further used as the prompt constraints for SAM to refine the anomaly segmentation results. In details, we introduce a crucial Unified Multi-scale Cross-modal Interaction (UMCI) module for interacting language with visual features at multiple scales of CLIP to reason anomaly positions. Then, we design a novel Multi-level Mask Refinement (MMR) module, which utilizes the positional information as multi-level prompts for SAM to acquire hierarchical levels of masks and merges them. Extensive experiments validate the effectiveness of our approach, achieving the optimal segmentation performance on the MVTec-AD and VisA datasets. Our code is public.\footnote{\url{https://github.com/Lszcoding/ClipSAM}}

%Recently, foundational models such as CLIP and SAM have demonstrated remarkable zero-shot capabilities. However, challenges persist in Zero-Shot Anomaly Segmentation (ZSAS). This is because CLIP classifies based on the similarity of global features from different modalities. It fails to align part-level features with language effectively, resulting in imprecise segmentation of anomalous parts. Meanwhile, although SAM exhibits remarkable segmentation capabilities, it lacks semantic understanding. Also, it generates numerous redundant masks without prompted constraints, leading to difficulties in post-processing. Therefore, considering the complementary strengths of both models, we propose a new framework named CLIP and SAM Collaboration. Our aim is to leverage CLIP's semantic information for anomaly localization and utilize it as a constraint prompt for SAM to refine our segmentation results. Specifically, we introduce a crucial Collaborative Row-and-Column Positioning (CRCP) module. It utilizes mechanisms for the interaction of language with visual row-and-column features respectively, collaboratively reasoning anomaly positions. Furthermore, we design a module named Multi-level Mask Refinement (MMR), which utilizes positional information as prompts for SAM to acquire masks at different hierarchical levels and merges them. Extensive experiments validate the effectiveness of our approach, achieving optimal segmentation performance on the MVTec-AD and VisA datasets.
\end{abstract}

% {\it IJCAI--24 Proceedings}    
\section{Introduction}
\label{sec:intro}
% 异常分割（AS）的目标是对物体异常部分进行定位和分割，这在工业质检中至关重要[   ]。现有的工作[   ]通常假设模型能在正常数据上进行训练，从而能够让模型认识正常并区别异常。然而，由于工业产品的多样性以及一些保密原则导致难以为每个类别收集大量的数据。这些方法在没有训练数据的类别上表现不佳。因此零样本异常分割（zsas）是一个具有挑战性的问题。

Zero-Shot Anomaly Segmentation (ZSAS) is a critical task in fields such as image analysis~\cite{fomalont1999image} and industrial quality inspection~\cite{vtadl,be,mvtec-ad}. Its objective is to accurately localize and segment anomalous regions within images, without relying on prior class-specific training samples. As a result, the diversity of industrial products and the uncertainty in anomaly types pose significant challenges for the ZSAS task.

% The goal of anomaly segmentation (AS) is to locate and segment abnormal parts of objects, which is crucial in industrial quality inspection ~\cite{vtadl,be,mvtec-ad}. 
% Existing works ~\cite{padim,cflow,unified,pro,fc} usually assumes that the model can be trained on a large number of normal data, and produce abnormal decision scores for anomalies. However, the diversity of industrial products and some confidential requirements make it difficult to collect a large amount of training data for each category, and these methods ~\cite{padim,cflow,unified,pro,fc}  will perform poorly on categories without training data, thus making zero-shot anomaly segmentation(ZSAS) challenging.

\begin{figure}[t]
  \centering
  \includegraphics[width=\linewidth]{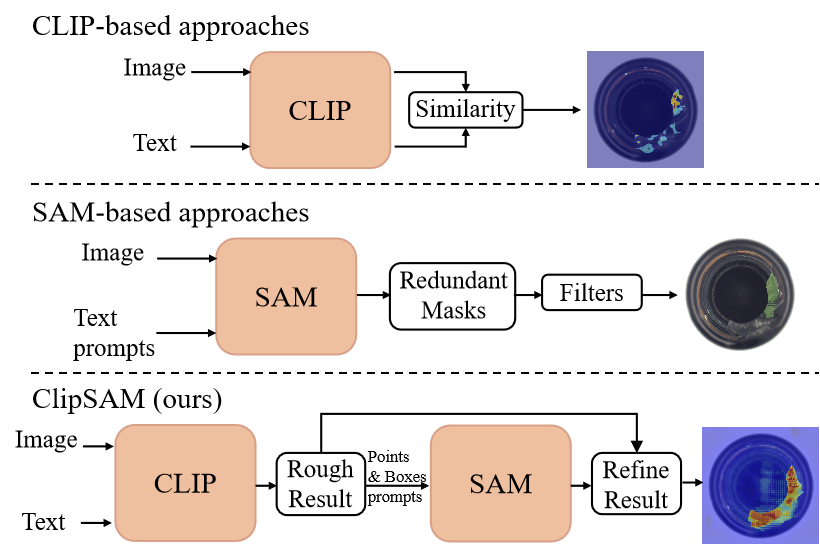}
  \vspace{-5mm}
  \caption{Structural comparisons among different approaches for Zero-Shot Anomaly Segmentation. Top: CLIP-based approaches. Middle: SAM-based approaches. Bottom: Our ClipSAM approach that leverages the strengths of both CLIP and SAM methods.}
  \label{fig:structure}
  \vspace{-5mm}
\end{figure}

% 近期的ZSAS工作,比如WinCLIP和APRIL-GAN，使用CLIP来对异常区域进行定位和分割。他们通过计算图像patch tokens和文本tokens之间的相似性来为每个patch进行是否异常的归类。然而，CLIP是为了对齐语言和视觉的全局类token。它的强语义理解能力并不能很好地作用于文本与patch tokens之间的对齐。而缺乏对于细粒度特征的理解会导致它对异常部分的定位和分割是粗糙的。
%此外，一些工作，比如SAA[]，尝试借鉴SAM强大的精确分割能力实现zsas。SAM能够接受不同的提示作为输入，比如点、框和文本来为分割结果提供指导。SAA采用词提示SAM以获取候选掩码并采用滤波器进行后处理。然而单个词对于物体part的位置描述弱于带有精确位置信息的空间提示比如点和框。
%基于此，我们首次提出CLIP与SAM协同解决ZSAS的框架，与先前工作的框架对比在图1中展示。我们希望借助CLIP的语义理解能力，实现对于异常部分的初步分割和定位，并根据定位信息生成相应的点和框作为提示指导SAM实现精准分割。

With the emergence of foundational models such as CLIP~\cite{clip} and SAM~\cite{sam}, the notable advancements have been achieved in Zero-Shot Anomaly Segmentation. 
% Recent works like WinCLIP~\cite{winclip} and APRIL-GAN~\cite{aprilgan} have employed CLIP to reason and identify anomalous regions.
As depicted in Figure~\ref{fig:structure}, the CLIP-based approaches, like WinCLIP~\cite{winclip} and APRIL-GAN~\cite{aprilgan}, determine the anomaly classification of each patch by comparing the similarity between image patch tokens and text tokens. 
While CLIP exhibits a strong semantic understanding capability, it is achieved by aligning the global features of language and vision, making it less suitable for fine-grained segmentation tasks~\cite{cris}. Due to the fact that anomalies consistently manifest in specific regions of objects, the global semantic consistency inherent in CLIP is unable to achieve precise identification of the edges of local anomalies.
% However, the CLIP model's strong semantic understanding, achieved through aligning global features of language and vision inputs, is not suitable for fine-grained segmentation tasks~\cite{cris}. 
On the other hand, researchers have explored the SAM-based approaches to assist the ZSAS task. SAM has superior segmentation capabilities and can accept diverse prompts, including points, boxes, and textual prompts, to guide the segmentation process. To this end, SAA~\cite{saa}, as shown in Figure~\ref{fig:structure}, utilizes the SAM with textual prompts to generate vast candidate masks and applies filters for post-processing.
However, simple textual prompts may be insufficient for accurately describing anomalous regions, resulting in subpar anomaly localization performance and underutilization of SAM's capabilities. Meanwhile, the ambiguous prompts lead to the generation of redundant masks, which requires further selection of correct masks.

Based on these observations, 
% and the proven effectiveness of combining foundational models~\cite{sam-clip}, 
we innovatively propose a CLIP and SAM collaboration framework, first employing CLIP for anomaly localization and rough segmentation and then utilizing SAM and the localization information to refine the anomaly segmentation results.
% our inquiry naturally arises: \emph{"Can the collaboration of CLIP and SAM enhance the efficiency of ZSAS?"}
% While previous CLIP-based methods~\cite{winclip,aprilgan} have shown promising results for ZSAS, their performance still has room for improvement due to the absence of cross-modal interaction.
% in anomaly classification, their limited ability to accurately localize anomalous regions can be attributed to the lack of cross-modal interaction. 
For the stage of CLIP, it is crucial to incorporate the fusion of language tokens and image patch tokens and model their dependencies to strengthen CLIP's ability to segment anomaly parts, since cross-modal interaction and fusion have been proven to be beneficial for localization and object segmentation in several studies~\cite{fuse1,fuse2,fuse3}.
% Recently, several studies~\cite{fuse1,fuse2,fuse3} have also shown the effectiveness of integrating features from different modalities for object localization and segmentation.
% Hence, it is crucial to incorporate the fusion of language tokens and image patch tokens and model their dependencies to strengthen CLIP's ability to segment anomaly parts.
Further, several notable works~\cite{acnet,ccnet,strip-pooling} have focused on enhancing the model's local semantic comprehension by paying attention to row and column features.
% integrating interactions between horizontal and vertical visual features with textual information.
Motivated by these studies, we have developed a novel cross-modal interaction strategy that facilitates the interaction of text and visual features at both row-column and multi-scale levels, adequately enhancing CLIP's capabilities for positioning and segmenting anomaly parts.
For the stage of SAM, in order to fully harness its fine-grained segmentation capability, we exploit the localization abilities of CLIP to provide more explicit prompts in the form of both points and bounding boxes. 
This approach greatly enhances SAM's ability to segment anomaly regions accurately.
Further, we have noticed that SAM's segmentation results often display masks with different levels of granularity, even when provided with the same prompt. 
% Traditional methods rely on intricate post-processing strategies to select masks with higher confidence, but these approaches are manually designed and may suffer from yielding inconsistent segmentation results. 
To avoid the inefficient post-processing caused by further mask filtering, we propose a more efficient mask refinement strategy that seamlessly integrates different levels of masks, leading to enhanced anomaly segmentation results.

% SAM's powerful part-level segmentation capabilities need to be constrained by precise prompts. 
% The disadvantage of the word prompt used by SAA is that it cannot specifically describe the abnormal area, resulting in poor segmentation prediction. 
% Therefore we use spatial prompts (points and boxes) with explicit location information. To this end, we further propose the Multi-level Mask Refinement module. This module looks for multi-level prompts based on CLIP's rough segmentation to guide SAM in refining the final result.

%Instead, we generate corresponding point and box prompts from positioning information. Accurate positioning prompts can constrain SAM to generate precise masks without the need for further filtering. Furthermore, we will integrate the masks with the preliminary results obtained from CLIP to achieve the final refined segmentation results.

% 在这项工作中，我们提出利用CLIP和SAM协同（）的新框架来进行ZSAS。（）分为两个部分。第一部分，我们利用CLIP来进行定位和粗分割。为了实现多模态特征在不同层级上的融合，我们设计Collaborative Row-and-Column Positioning（CRCP）模块。具体来说，CRCP从水平和垂直两个方向将image patch tokens聚合成对应的行级和列级特征。让它们与文本交互以在不同方向上感知局部异常。同时CRCP还考虑多尺度的全局视觉语义，以在不同层级上认识异常。第二部分，我们利用SAM来细化结果。为此我们设计了Multi-level Mask Refinement（MMR）模块。具体来说，我们从发CLIP的粗分割结果上生成指示异常位置的空间提示从而引导SAM生成精确掩码。最后我们基于不同的掩码置信度将它们与CLIP得到的结果融合。
As a conclusion, we propose a novel two-stage framework named CLIP and SAM Collaboration (ClipSAM) for ZSAS. The structural comparisons with previous works are illustrated in Figure~\ref{fig:structure}. In the first stage, we employ CLIP for localization and rough segmentation. To achieve the fusion of multi-modal features at different levels, we design the Unified Multi-scale Cross-modal Interaction (UMCI) module. UMCI aggregates image patch tokens from both horizontal and vertical directions and utilizes the corresponding row and column features to interact with language features to perceive local anomalies in different directions.
% into corresponding row and column features. This allows them to interact with language tokens to perceive local anomalies in different directions. 
UMCI also considers the interaction of language and multi-scale visual features.  
% to recognize anomalies at different levels. 
In the second stage, we exploit the CLIP's localization information to guide SAM for segmentation refinement. Specifically, we propose the Multi-level Mask Refinement (MMR) module, which first extracts diverse point and bounding box prompts from the CLIP's anomaly localization results and then uses these prompts to guide SAM to generate precise masks. 
% For this purpose, we design locations based on the results of CLIP, guiding SAM to generate precise masks. 
Finally, we fuse these masks with the results obtained from CLIP based on different mask confidences.
% 为了利用不同大模型的特性来更好地实现ZSAS，我们首次提出CLIP和SAM协同进行异常分割的框架。具体地说，我们先用CLIP得到一个（粗分割结果和定位信息），再利用SAM和定位提示来细化结果。
% 为了帮助CLIP实现推理定位，我们提出了Collaborative Row-and-Column Positioning (CRCP) module。通过文本特征与不同层级的视觉特征交互以学习关于异常部分的局部和全局语义。并为SAM的分割提供精确的定位引导。
% 为了用SAM细化分割结果，我们设计了Multi-level Mask Refinement（MMR）模块。它能基于CLIP对目标的定位来引导SAM生成精准的掩码，并将其与CLIP的结果融合，实现精细的分割。
% 在不同的数据集上进行的广泛实验一致地验证了我们方法可以实现最先进的零样本分割性能。例如，特别是在MVTec-AD数据集的像素级AUROC、F1-max和Pro的指标上，我们的方法+15.2↑/+11.7↑/+28.5↑超过SOTA WinCLIP。
Our main contributions can be summarized as follows:
\begin{itemize}
    \item We propose a novel framework named CLIP and SAM Collaboration (ClipSAM) to fully leverage the characteristics of different large models for ZSAS. Specifically, we first use CLIP to locate and roughly segment the anomaly objects, and then refine the segmentation results with SAM and the positioning information.
    % \item We propose a novel framework named CLIP and SAM Collaboration (ClipSAM) to fully leverage the characteristics of different large models for ZSAS. Specifically, we first use CLIP to realize localization and rough segmentation, and then refine the segmentation results with SAM and the localization information.
    % obtain the localization information and rough segmentation results, and then refine the results with SAM and positional prompts.
    \item To better assist CLIP in realizing desired localization and rough segmentation, we propose the Unified Multi-scale Cross-modal Interaction (UMCI) module, which learns local and global semantics about anomalous parts by interacting language features with visual features at both row-column and multi-scale levels.
    % different hierarchical levels.
    % and provides precise positioning guidance for the segmentation of SAM.
    \item To refine the segmentation results with SAM adequately, we designed the Multi-level Mask Refinement (MMR) module. It extracts point and bounding box prompts from the CLIP's localization information to guide SAM in generating accurate masks, and fuse them with the results of CLIP to achieve fine-grained segmentation.
    \item Extensive experiments on various datasets consistently validate that our approach can achieve new state-of-the-art zero-shot anomaly segmentation results. Particularly on the MVTec-AD dataset, our method outperforms the SAM-based method by +19.1↑ in pixel-level AUROC, +10.0↑ in $F_{1}$-max and +45.5↑ in Pro metrics.
\end{itemize}
\section{Related Work}
\begin{figure*}[t]
  \centering
  \includegraphics[width=0.96\linewidth]{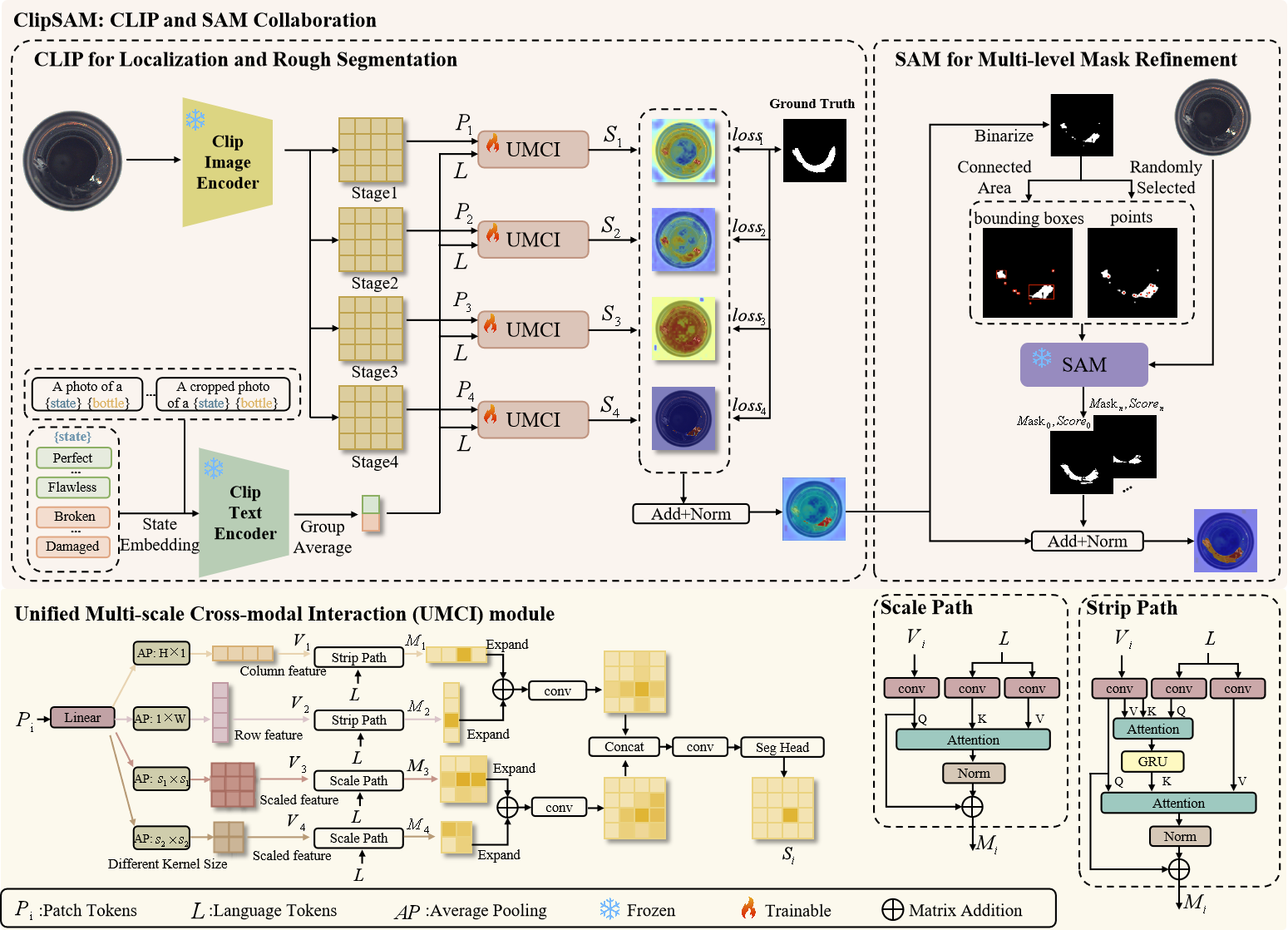}
  \vspace{-2mm}
  \caption{Overview of the proposed ClipSAM framework. ClipSAM includes two main processes: using CLIP for localization and rough segmentation, and using positioning information to prompt SAM to refine the segmentation results. These processes contain two important components: the Unified Multi-scale Cross-modal Interaction (UMCI) module and the Multi-level Mask Refinement (MMR) module. The UMCI module is employed for the interaction of language features with visual features of different directions and scales, facilitating CLIP's ability to locate and segment anomaly objects. Meanwhile, the MMR module combines SAM, and uses point and box prompts extracted from location information to guide SAM to output the desired masks, and fuses them with the rough segmentation result obtained by CLIP.
  }
  \vspace{-2mm}
  \label{fig:framework}
\end{figure*}

% 它包括两个主要流程：用CLIP进行定位和粗分割，以及用定位信息提示SAM来细化分割结果。这其中包括两个重要组件：协同行列定位（CRCP）模块和多层级掩码细化（MMR）模块。CRCP模块被用来进行语言特征和不同方向，不同尺度的视觉特征的交互，促进了CLIP对物体部分的定位和分割能力。同时，MMR模块结合了SAM模型。它使用具有位置信息的空间提示引导SAM输出期望的掩码并将其与CLIP获得的粗分割结果融合。

\subsection{Zero-shot Anomaly Segmentation}
The methods for Zero-Shot Anomaly Segmentation (ZSAS) can be mainly divided into two categories. The first category is based on CLIP. As the pioneering work, WinCLIP~\cite{winclip} calculates the similarity between image patch tokens and textual features for ZSAS. Further, APRIL-GAN~\cite{aprilgan} employs linear layers to better align features of different modalities. AnoVL~\cite{AnoVL} and ANOMALYCLIP~\cite{Anomalyclip} propose to enhance the generalization of text. SDP~\cite{sdp} proposes to address noise in the encoding process of the CLIP image encoder. 
The second category is based on SAM. Specifically, SAA~\cite{saa} utilizes text prompts for SAM to generate vast candidate masks and uses a complex evaluation mechanism to filter out irrelevant masks.

% Recently, researchers have explored the utilization of CLIP and SAM for zero-shot anomaly segmentation. 
However, relying solely on CLIP or SAM may lead to certain limitations. For instance, CLIP-based methods struggle with precise segmentation of local anomalies, while SAM-based methods heavily rely on specific prompts. To address these drawbacks, we explore the collaboration mechanism of CLIP and SAM and propose a novel framework to leverage their strengths. Besides, we further design the Unified Multi-scale Cross-modal Interaction (UMCI) module and Multi-level Mask Refinement module to better exploit the specific ability of CLIP and SAM, respectively.

\subsection{Foundation Models}
\label{sec:model}
% In recent years, models are trained with more and more data to achieve strong generalization. CLIP~\cite{clip} and SAM~\cite{sam} are two representative models and have demonstrated amazing zero-shot reasoning capabilities in classification and segmentation tasks respectively. CLIP is dedicated to the alignment of multi-modal features and has robust semantic understanding capabilities for language and vision. SAM can easily achieve fine-grained segmentation based on different prompts. Recently, researchers try to use CLIP and SAM for zero-shot anomaly segmentation. However, using them alone can be problematic. For example, CLIP cannot achieve precise location and segmentation of local anomalies. SAM relies on specific prompts. Therefore, we believe it is essential to use CLIP to provide guidance for SAM, leveraging the advantages of both in a zero-shot setting.
Recently, there has been an increasing focus and attention on foundation models. Various foundation models have achieved satisfactory performance on kinds of downstream tasks~\cite{bert,gpt3}. Notably, CLIP~\cite{clip} and SAM~\cite{sam} have emerged as two representative models with impressive zero-shot reasoning capabilities in classification and segmentation tasks, respectively. More specifically, CLIP focuses on aligning multi-modal features and possesses robust semantic understanding abilities for both language and vision, while SAM excels in achieving fine-grained segmentation based on different prompts. 
Recently,~\cite{sp-sam} attempts to establish a connection between SAM's image encoder and CLIP's text encoder for surgical instrument segmentation,~\cite{sam-clip} attempts to merge SAM and CLIP to facilitate downstream tasks. 
These works highlight the critical importance of exploring collaboration among foundational models.

\subsection{Cross-modal Interaction}
\label{sec:ci}
% With the development of visual-language models, how to utilize cross-modal information for segmentation tasks has become a challenge.
% In the field of multi-modal learning, cross-modal interaction is becoming increasingly important. This method ~\cite{fuse5} concatenates features of different modalities and uses convolution to fuse. STEP ~\cite{fuse5} establish correlation between important areas in the image and keywords in the text to improve the fusion of cross-modal information. BRINet~\cite{fuse3} exchanges cross-modal information between different blocks of the encoder to facilitate image segmentation. Cross-modal interaction has been successful in different domains, which motivated us to explore this issue in ZSAS. The Unified Multi-scale Cross-modal Interaction module is defined to effectively handles the problem that the abnormal area is usually located in the local position of the object. It considers both the interaction of text with row- and column-level visual features and multi-scaled visual features.
In the field of multi-modal learning, cross-modal interaction is becoming increasingly important. Specifically,~\cite{fuse4} concatenates features from different modalities and uses convolutions for multi-modal information fusion. Further, STEP~\cite{fuse5} establishes correlations between important areas in the image and relevant keywords in the text to enhance the fusion of cross-modal information. BRINet~\cite{fuse3} exchanges cross-modal information between different blocks of the encoder to facilitate image segmentation. The success of cross-modal interaction in diverse domains has motivated us to explore it in the context of zero-shot anomaly segmentation. To effectively address the challenge of localizing abnormal regions within an object, we introduce the Unified Multi-scale Cross-modal Interaction module, taking into account the interaction between text and both row-column and multi-scaled visual features.

\section{Methodology}
\subsection{CLIP and SAM Collaboration}
\label{sec:clipsam}
CLIP has a strong semantic understanding of different modalities, while SAM can easily detect edges of fine-grained objects, both of which are important for anomaly segmentation. 
% As illustrated in Sec.~\ref{sec:intro}, although each of them is limited by the model itself, ~\cite{sp-sam,sam-clip} have begun to explore building knowledge bridges between them to collaboratively assist downstream tasks.
In this paper, we present a novel CLIP and SAM collaboration framework called ClipSAM, which aims to boost the performance of ZSAS. The overview architecture is illustrated in Fig.~\ref{fig:framework}. Specifically, we leverage the CLIP for initial rough segmentation and utilize it as a constraint to refine the segmentation results with SAM.
In Sec.~\ref{sec:umci}, we introduce the Unified Multi-scale Cross-modal Interaction (UMCI) module within the CLIP stage to achieve accurate rough segmentation and anomaly localization. In Sec.~\ref{sec:mmr}, we design the Multi-level Mask Refinement (MMR) module, which incorporates the guidance from CLIP to facilitate SAM in generating more precise masks for achieving fine-grained segmentation. In addition, the optimization function of the overall framework is discussed in Sec.~\ref{sec:of}.

\subsection{Unified Multi-scale Cross-modal Interaction}
% \subsection{CLIP with Unified Multi-scale Cross-modal Interaction}
\label{sec:umci}
%为了对不同模态的特征进行深层次的融合，实现对于局部异常的定位。我们提出Collaborative Row-and-Column Positioning （CRCP）模块来对其位置进行推理。具体的说，如图5所示，CRCP包含了两条并行的路径，分为条带路径和尺度路径。一方面我们利用条带路径来realizes the positioning target by perceiving the row- and column-wise local features of the image.另一方面我们利用尺度路径来感知图像的全局特征来实现对于目标物体的分割。最后我们利用卷积层实现对于两条路径知识的融合。
In our ClipSAM framework, the CLIP encoder is employed to process both text and image inputs. For a specific pair of text and image, the encoder generates two outputs: $L\in \mathbb{R}^{C_{t}\times 2}$ and $P_{i}\in \mathbb{R}^{H\times W\times C}$. Here, $L$ represents the textual feature vector, in line with WinCLIP \cite{winclip}, reflecting two categories of normal and abnormal. $P_{i}$ denotes the patch tokens derived from the i-th stage of the encoder. For more details of the textual feature $L$, please refer to \textbf{Appendix A}.
% Given a text and image input, the corresponding CLIP encoder can be utilized to obtain $L\in \mathbb{R}^{C_{t}\times 2}$ and $P_{i}\in  \mathbb{R}^{H\times W\times C}$, where $L$ is consistent with the design of text in WinCLIP ~\cite{winclip}, 2 represents the two categories of normal and abnormal and $P_{i}$ is the patch tokens of the i-th stage of ViT. Please refer to Appendix A for details about the texts.

% To achieve deep integration of features from different modalities for the purpose of local anomaly localization, we propose the Unified Multi-scale Cross-modal Interaction (UMCI) module to infer their positions. Specifically, as shown in Figure~\ref{fig:framework}, UMCI comprises two parallel paths, 

As discussed in Sec.~\ref{sec:intro}, cross-modal fusion has been proven to be beneficial for object segmentation.
% With the textual feature vector $L$ and patch tokens $P_{i}$ of various scales derived from the encoder, 
To achieve this, the Unified Multi-scale Cross-modal Interaction (UMCI) module is designed. 
% to infer abnormal locations. 
Specifically, the UMCI model consists of two parallel paths: the ${Strip}$ ${Path}$ and the ${Scale}$ ${Path}$. The ${Strip}$ ${Path}$ 
% is designed for anomaly localization, capturing 
captures both row- and column-level features of the patch tokens to precisely pinpoint the location. 
% In contrast, the 
The ${Scale}$ ${Path}$ focuses on grasping the image's global features of various scales, enabling a comprehensive understanding of the anomaly.
% 's overall semantics. 
More details are described below.
% The details and fusion process of these two paths are described below.
% As shown in Figure~\ref{fig:framework}, We use CLIP to reason abnormal locations through different modal interactions. For this purpose, we design the Unified Multi-scale Cross-modal Interaction (UMCI) module, which comprises two parallel paths called the ${Strip}$ ${Path}$ and the ${Scale}$ ${Path}$. On one hand, we utilize the ${Strip}$ ${Path}$ to achieve localization by perceiving the row- and column-level features of the image. On the other hand, we use the ${Scale}$ ${Path}$ to perceive the global features of the image for understanding the overall semantics of the anomaly. 
% Finally, we employ convolutional layers to achieve the fusion of knowledge from the two paths.
%在条带路径中，如图所示，它首先将视觉特征沿着水平和垂直两个方向分解成行级和列级两个部分。具体的说，他接受L和P作为输入，这里我们为了便于表示，去掉了描述不同stage的下标i。首先，利用两个核大小分别为${1\timesW}$和${H\times1}$的平均池化层，然后使用${1\times1}$的卷积层和Gelu激活函数。其中AP表示平均池化，v_{row}\in \mathbb{R}^{H \times C_{h}},v_{column}\in \mathbb{R}^{W \times C_{w}}。由于后续对于行列特征的计算相似，我们以行特征为例。我们对文本特征使用卷积层${1\times1}$的卷积层It first decomposes visual features along horizontal and vertical directions into two components: row- and column-level features. Specifically,

\noindent
(1) \textbf{${Strip}$ ${Path}$}. 
Denote the inputs of a specific UMCI module as textual feature vector $L$ and patch tokens ${P}$.
We first process the patch tokens to grasp the visual features.
The image features are projected to align with the text features in dimension, resulting in $\hat{P}\in \mathbb{R}^{H\times W\times C_{t}}$. 
To extract row- and column-level features from $\hat{P}$, we apply two average pooling layers, with kernel sizes of ${1\times W}$ and ${H\times 1}$ respectively:
\begin{equation}
\begin{aligned}
    &v_{row} = conv_{1\times 3} (Avg\_Pool_{1\times W} (\hat{P})), \\
    &v_{col} = conv_{3\times 1} (Avg\_Pool_{H\times 1} (\hat{P})),
\end{aligned}
\end{equation}
where $v_{row}\in \mathbb{R}^{H \times c_{h}}$ and $v_{col}\in \mathbb{R}^{W \times c_{h}}$ are the row-level and column-level features. ${H}$ and ${W}$ denote the height of the vertical feature and the width of the horizontal feature.

We then focus on the internal process of the ${Strip}$ ${Path}$. Take $v_{row}$ for example, we apply convolution layers to the text features $L$, obtaining $t_{row}^{1}, t_{row}^{2}\in \mathbb{R}^{c_{h} \times 2}$. $t_{row}^{1}$ and $t_{row}^{2}$ serve as the text feature input of the subsequent Scaled Dot-Product Attention mechanism~\cite{transformer}, a crucial component for the interaction between the language and visual domains. Specifically, we implement a two-step attention mechanism to efficiently predict the language perception of the pixels in $v_{row}$ (normal or abnormal), denoted as $M_{row}\mathbb{R}^{H \times c_{h}}$. 
% Detailed analysis and equations 
More details are provided in \textbf{Appendix B}. 
In parallel, $v_{col}$ is similarly processed to obtain $M_{col}\in \mathbb{R}^{W \times c_{h}}$.

We then utilize the bilinear interpolation to expand $M_{row}$ and $M_{col}$ to their original scales and combine the results:
\begin{equation}
\begin{aligned}
    &M_{row,col} = conv_{3\times 3}(B(M_{row})+B(M_{col})),
\end{aligned}
\end{equation}
where ${B}$ symbolizes the bilinear interpolation layer. This process results in  $M_{row,col}\in \mathbb{R}^{H\times W\times c_{h}}$.

\noindent
(2) \textbf{${Scale}$ ${Path}$}. 
In this path, the image features are also projected to $\hat{P}\in \mathbb{R}^{H\times W\times C_{t}}$. Then we apply two average pooling layers with kernel sizes of $s_1$ and $s_2$ to grasp the visual features of different scales:
\begin{equation}
\begin{aligned}
    &v_{{g}_{1}} = conv_{3\times 3}^{{g}_{1}} (Avg\_Pool_{{s}_{1}\times {s}_{1}} (\hat{P})), \\
    &v_{{g}_{2}} = conv_{3\times 3}^{{g}_{2}} (Avg\_Pool_{{s}_{2}\times {s}_{2}} (\hat{P})),
\end{aligned}
\end{equation}
where $v_{{g}{1}}\in \mathbb{R}^{h{{g}{1}}\times w{{g}{1}}\times c{h}}$ and $v_{{g}{2}}\in \mathbb{R}^{h{{g}{2}}\times w{{g}{2}}\times c{h}}$ represent visual features at different scales.

For the internal process of ${Scale}$ ${Path}$, we consider $v_{{g}_{1}}$ for example. The text features are processed by convolution layers and yield $t_{{g}_{1}}^{k}, t_{{g}_{1}}^{v}\in \mathbb{R}^{c_{{g}_{1}} \times 2}$. We then obtain the language perception of the pixels $M_{{g}_{1}}\in \mathbb{R}^{h_{{g}_{1}}\times w_{{g}_{1}}\times c_{h}}$ by the attention with $v_{{g}_{1}}$ as the query,  $t_{{g}_{1}}^{k}$ as the key and  $t_{{g}_{1}}^{v}$ as the value.
More details are provided in \textbf{Appendix B}. Similar to the ${Strip}$ ${Path}$, we utilize the bilinear interpolation to resize and combine $M_{{g}_{1}}$ and $M_{{g}_{2}}$:
\begin{equation}
\begin{aligned}
    &M_{{g}_{1},{g}_{2}} = conv_{3\times 3}^{{g}_{1},{g}_{2}}(B(M_{{g}_{1}})+B(M_{{g}_{2}})),
\end{aligned}
\end{equation}
\noindent
(3) \textbf{${Dual-path}$ ${Fusion}$.}
After the ${Strip}$ ${Path}$ and ${Scale}$ ${Path}$, we have obtained the pixel-wise predictions $M_{row,col}$ and $M_{{g}_{1},{g}_{2}}$, providing comprehensive location and semantics information about the anomaly. The last step of the UMCI module involves fusing these results to get the rough segmentation of the anomalous region. Specifically, we introduce a residual connection from the input patch token $\hat{P}$, and fuse it with the pixel-wise predictions by a convolution layer:
\begin{equation}
\begin{aligned}
    &v_{ori} = conv_{3\times 3}^{ori}(\hat{P}),\\
    &M_{all}= conv_{3\times 3}^{all}(concat(v_{ori},M_{row,col},M_{{g}_{1},{g}_{2}})).\\
    % &O = MLP(ReLU(M_{all}+\hat{P})),
\end{aligned}
\end{equation}
We employ a Multi-Layer Perceptron as the segmentation head, and the rough segmentation of the anomalous regions is mathematically described as: 
\begin{equation}
O = MLP(ReLU(M_{all}+\hat{P})).
\end{equation}
where $O\in \mathbb{R}^{H\times W\times 2}$ denotes the segmentation output of a specific UMCI module, and dimension 2 represents the classification of the foreground anomalous parts and the background.
Assume there are $n$ stages in the encoder, and denote $O_i$ as the segmentation output of stage ${i}$. Then the final segmentation results can be calculated as $O=\frac{1}{n}\textstyle\sum_{i=1}^{n} O_{i}$.

\iffalse
We merge the results of ${Strip}$ ${Path}$ and ${Scale}$ ${Path}$. Specifically, a convolutional layer is applied to ${\hat{P}}$ and fuse it with $M_{row,col}$ and $M_{{g}_{1},{g}_{2}}$. Also, ${\hat{P}}$ is added to the fused features through residual connection. Finally, we use an MLP as our segmentation head to perform rough segmentation of anomalous region:

\begin{equation}
\begin{aligned}
    &v_{ori} = conv_{3\times 3}^{ori}(\hat{P}),\\
    &M_{all}= conv_{3\times 3}^{all}(concat(v_{ori},M_{row,col},M_{{g}_{1},{g}_{2}})),\\
    &O = MLP(ReLU(M_{all}+\hat{P})),
\end{aligned}
\end{equation}
where $O\in \mathbb{R}^{H\times W\times 2}$ and ${2}$ represents foreground that contains abnormal parts and background.

Suppose we have n stages in total. $O_{i}$ can be got by UMCI For different stages ${i}$. The rough segmentation result based on CLIP is the summation of these results:
\begin{equation}
\begin{aligned}
    &O=\frac{1}{n}\textstyle\sum_{i=1}^{n} O_{i}
\end{aligned}
\end{equation}
%对于不同的stage i，我们能用CRCP得到i个结果，我们基于CLIP的粗分割结果是对这i个结果的求和：
\fi

%CRCP模块能够通过模态间的交互实现对于异常部分的定位和粗分割，我们需要借助SAM对于part级精确分割的能力对于结果进行细化。因此我们提出Multi-level Mask Refinement（MMR）模块。如sec.1中讨论的那样，SAM的精准分割能力需要借助具体的关于位置的提示。我们在这里选择空间提示，也就是点和框来为SAM提供引导。具体地说，如图所示。MMR接受粗分割结果O作为输入，由于我们这里将异常部分和其他区域划分为前景和背景，所以我们取前景作为我们后处理的输入O。我们首先根据阈值对其进行二值化。是针对于检测结果的一个二值化掩码，我们从他的连通区域中寻找点和框，具体的，对于点，我们在连通区域中随机选取十个点。对于，我们利用现有的函数根据二值掩码的连通区域的大小得到框的信息, 对于第i个框，我们会得到Sbi=【（xbi）,(ybi),(hbi),(wbi)】,如图所示，单独使用点或框也有可能导致结果的偏差。相反，我们用点和框相互约束和补充，能够更精确的给出细粒度分割结果。
\subsection{Multi-level Mask Refinement}

\label{sec:mmr}
\begin{figure}[t]
  \centering
  \includegraphics[width=\linewidth]{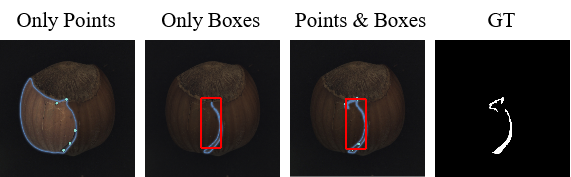}
  \vspace{-7mm}
  \caption{The results produced by SAM with different spatial prompts. As we can see, constraining SAM with the spatial prompt that represents points and boxes as a whole leads to better results.
  % constraining SAM with prompts of points and boxes as a whole gives better results.
  }
  \label{fig:pointandbox}
  \vspace{-2mm}
\end{figure}
% The conventional direct application of SAM in the ZSAS task is unsatisfactory. 
With the rough segmentation $O$ from the CLIP phase, we propose the Multi-level Mask Refinement (MMR) module to extract point and box prompts to guide SAM to generate accurate masks.
% and fine-grained segmentation results. 
In the MMR module, the foreground of the rough segmentation, denoted as $O_{f}\in \mathbb{R}^{H\times W}$, is firstly post-processed with a binarization step to obtain a binary mask ${O_{b}(x,y)}$. Denote $v(x, y)_f, x\in{H},y\in{W}$ as the value of a specific pixel in $O_{f}$, then the value of each pixel $v(x, y)_b$ in ${O_{b}(x,y)}$ can be calculated as:
\begin{equation}
\begin{aligned}
    v(x, y)_b=\begin{cases}
  & 1,\text{if  }  v(x, y)_f > threshold \\
  & 0, otherwise
\end{cases} 
\end{aligned}
\end{equation}
where $threshold$ represents the binary threshold, and the value 1 corresponds to the anomalous pixels. 
Within the connected areas of this binary mask, we identify some boxes and points to provide spatial prompts for SAM. For point selection, $m$ random points are chosen, represented as $S_{p}=[(x_{p_{1}},y_{p_{1}}),\dots,(x_{p_{m}},y_{p_{m}})]$, where $(x_{p_{i}},y_{p_{i}})$ represents the position of the i-th point. 
Boxes are generated based on the size of connected regions in the binary mask, with the i-th box denoted by $S_{b_{i}}=[(x_{b_{i}},y_{b_{i}},h_{b_{i}},w_{b_{i}})]$. The complete set with $q$ boxes is represented as $S_{b}=[S_{b_{1}},\dots,S_{b_{q}}]$. 

With the point prompts $S_{p}$ and box prompts $S_{b}$, we explore the optimal prompt sets for SAM. As can be seen in Figure~\ref{fig:pointandbox}, employing either points or boxes alone leads to biased results. In contrast, the combined application of both points and boxes can yield more precise and detailed segmentation. Therefore, in our ClipSAM framework, we use $S=S_{b}\cup{S_{p}}$ as the prompt set for SAM. With the original image $I$ and spatial prompts $S$ as inputs, SAM generates encoded features ${z}_{i}$ and ${z}_{s}$. The decoder within SAM then outputs the refined masks and corresponding confidence scores:
\begin{equation}
\begin{aligned}
    &( {masks},{scores}  ) =  \mathbb{D}^{sam} ( {z}_{i}\mid {z}_{s}  ).
\end{aligned}
\end{equation}
Each box shares the same point constraints, resulting in $q$ distinct segmentation masks. In our ClipSAM framework, SAM is configured to produce three masks with varying confidence scores for each box, represented as $masks=[(m_{1}^{1},m_{1}^{2},m_{1}^{3});\dots;(m_{q}^{1},m_{q}^{2},m_{q}^{3})]$ and $scores=[(s_{1}^{1},s_{1}^{2},s_{1}^{3});\dots;(s_{q}^{1},s_{q}^{2},s_{q}^{3})]$. 
The final fine-grained segmentation result $O_{final}$ is obtained by normalizing the fusion of rough segmentation and the refined masks:
\begin{equation}
\begin{aligned}
    &O_{final}=Norm(O+\sum_{i=1}^{q} \sum_{j=1}^{3}m_{i}^{j} \times s_{i}^{j}).
\end{aligned}
\end{equation}

\subsection{Objective Function}
\label{sec:of}
In our ClipSAM framework, the only part involving training is the UMCI module. To effectively optimize this module, we employ the Focal Loss \cite{focal} and the Dice Loss \cite{dice}, both of which are well-suited for segmentation tasks.

\noindent
\textbf{Focal Loss.} 
Focal loss is primarily applied to address class imbalance problem, a common challenge in segmentation tasks. It is appropriate for anomaly segmentation because, usually, the anomaly merely occupies a small fraction of the entire object. The expression of Focal loss is:
\begin{equation}
\begin{aligned}
    &l_{focal}=-\frac{1}{H\times W}\textstyle \sum_{i=0}^{H\times W}(1-p_{i})^{\gamma }\log_{}{(p_{i})},
\end{aligned}
\end{equation}
where $p_{i}$ is the predicted probability for a pixel being abnormal, and $\gamma$ is a tunable parameter and set to 2 in our paper.

\noindent
\textbf{Dice Loss.}
Dice Loss calculates a score based on the overlap between the target area and the model's output. This metric is also effective for class imbalance issue. Dice Loss can be calculated as:
\begin{equation}
\begin{aligned}
    &l_{dice}=1-\frac{1}{N}\frac{ 2\times {\textstyle \sum_{i=1}^{N}} y_{i}\hat{y}_{i} }{ {\textstyle \sum_{i=1}^{N}} y_{i}^{2}+{\textstyle \sum_{i=1}^{N}}\hat{y}_{i}^{2}}  ,
\end{aligned}
\end{equation}
where $N=H\times W$ is the total number of pixels in features.

\noindent
\textbf{Total Loss.}
We set separate loss weights for each stage and the total loss can be expressed as:
\begin{equation}
\begin{aligned}
    &l_{all} = \textstyle \sum_{i=1}^{n}\lambda_{i}(l_{focal}^{i}+l_{dice}^{i}),
\end{aligned}
\end{equation}
where $i$ denotes the index of stages, $\lambda_{i}$ is the loss weight of the i-th stage. The CLIP encoder in our implementation consists of 4 stages in total, and we set the loss weights for these stages at $0.1$, $0.1$, $0.1$, and $0.7$ respectively.

\section{Experiments}
\begin{table*}[!htbp]
\small
% \Huge
\centering
\resizebox{0.9\linewidth}{!}{
\begin{tabular}{l|l|cccc|cccc}
\toprule
&  &    \multicolumn{4}{c|}{MVTec-AD}& \multicolumn{4}{c}{VisA}  \\ 
\multirow{-2}{*}{Base model}  & \multirow{-2}{*}{Method}  & \multicolumn{1}{c}{AUROC}   & \multicolumn{1}{c}{$F_{1}$-max}   & \multicolumn{1}{c}{AP} & \multicolumn{1}{c|}{PRO}   & \multicolumn{1}{c}{AUROC}   & \multicolumn{1}{c}{$F_{1}$-max}   & \multicolumn{1}{c}{AP} & \multicolumn{1}{c}{PRO}          \\ 

% \multirow{2}*{Base model} & \multirow{2}*{Method} & \multicolumn{4}{c|}{MVTec-AD}  & \multicolumn{4}{c}{VisA} \\
% \cmidrule[0.6pt](r){3-6} \cmidrule[0.6pt](r){7-10}
% & & AUROC & $F_{1}$-max & AP & PRO & AUROC & $F_{1}$-max & AP & PRO \\
\midrule
 \multirow{4}{*}{\makecell{CLIP-based \\ Approaches}} & WinCLIP & 85.1 & 31.7 & - & 64.6 & 79.6 & 14.8 & - & 56.8 \\
 & APRIL-GAN & 87.6 & 43.3 & 40.8 & 44.0 & 94.2 & 32.3 & 25.7 & 86.8 \\
 & SDP & 88.7 & 35.3 & 28.5 & 79.1 & 84.1 & 16.0 & 9.6 & 63.4 \\
 & SDP+ & 91.2 & 41.9 & 39.4 & 85.6 & 94.8 & 26.5 & 20.3 & 85.3 \\ \midrule
 \multirow{2}{*}{\makecell{SAM-based \\ Approaches}}& SAA & 67.7 & 23.8 & 15.2 & 31.9 & 83.7 & 12.8 & 5.5 & 41.9 \\
 & SAA+ & 73.2 & 37.8 & 28.8 & 42.8 & 74.0 & 27.1 & 22.4 & 36.8 \\
\midrule
 CLIP \& SAM& \textbf{ClipSAM(Ours)} & \textbf{92.3} & \textbf{47.8} & \textbf{45.9} & \textbf{88.3} & \textbf{95.6} & \textbf{33.1} & \textbf{26.0} & \textbf{87.5} \\
\bottomrule
\end{tabular}
}
\vspace{-2mm}
\caption{Performance comparison of different kinds of ZSAS approaches on the MVTec-AD and VisA datasets. Evaluation metrics include AUROC, $F_{1}$-max, AP, and PRO. 
% under zero-shot setting. 
Bold indicates the best results.}
\vspace{-2mm}
\label{tab: Quantitative Results}
\end{table*}

\subsection{Experimental Setup}

\textbf{Datasets.} 
In this study, we conduct experiments on two commonly-used datasets of industrial anomaly detection, namely VisA \cite{visa} and MVTec-AD \cite{mvtec-ad}, which encompass a diverse range of industrial objects categorized as normal or abnormal. 
We follow the same training setup as existing zero-shot anomaly segmentation studies~\cite{winclip,aprilgan} to evaluate the performance of our method. Specifically, the model is first trained on the MVTec-AD dataset and then tested on the VisA dataset, and vice versa.
Additional experimental results on other datasets are provided in \textbf{Appendix C}.
% Our experiments are based on VisA~\cite{visa} and MVTec-AD~\cite{mvtec-ad} datasets. They all contain different objects in the industrial field and each category has normal samples and abnormal samples. They are commonly used data sets in the field of industrial anomaly detection. Our zero-shot anomaly segmentation task is set up in the same manner as previous works~\cite{winclip,aprilgan}. Specifically, when testing on MVTec-AD dataset, ZSAS allows training on any other dataset, where we use VisA dataset to train UMCI, and vice versa. This training approach demonstrates the generalization of our model. For experimental results on additional datasets, please refer to Appendix C.
%我们的实验基于MVTec-AD和VisA数据集。他们都包含了工业领域上的不同对象，每个类别都有正常数据和异常数据。他们是工业缺陷检测领域中常用的数据集。我们的零样本异常分割任务设置采用与之前工作一样的方式。具体地说，对于数据集MVTec-AD，ZSAS允许在除了这个数据集之外的任何数据集上进行训练。于是在测试MVTec-AD时我们选择用VisA数据集进行训练。反之亦然。这样的训练方式展现了我们模型的泛化性，在更多数据集上的实验结果请参阅Appendix A。

\noindent
\textbf{Metrics.} 
Following~\cite{winclip}, we employ widely-used metrics, i.e., AUROC, AP, $F_1$-max, and PRO~\cite{PROm}, to provide a fair and comprehensive comparison with existing ZSAS methods.
Specifically, AUROC reflects the model's ability to distinguish between classes at various threshold levels. AP quantifies the model's accuracy across different levels of recall. $F_1$-max is the harmonic mean of precision and recall at the optimal threshold, implying the accuracy and coverage of the model. PRO assesses the proportion of correctly predicted pixels within each connected anomalous region, offering insights into the model's local prediction accuracy.
Higher values of these metrics mean better performance of the evaluated method.
% The previous anomaly segmentation tasks~\cite{winclip} commonly utilized AUROC, AP, F1-max and PRO as evaluation metrics. % AUROC, AP and PRO are calculated based on whether each pixel is correctly classified. PRO~\cite{PROm} specifically calculates the proportion of anomalous pixels successfully detected within each connected anomalous component. Higher values for all these metrics indicate better model performance. 

\noindent
\textbf{Implementation details.} 
In the experiments, the pre-trained ViT-L-14-336 model released by OpenAI, which consists of 24 Transformer layers, is utilized for CLIP encoders. We extracted the image patch tokens after each stage of the image encoder (i.e., layers 6, 12,18, and 24) for the training of our proposed UMCI module, respectively. The optimization process is conducted on a single NVIDIA 3090 GPU using AdamW optimizer with the learning rate of ${1\times 10^{-4}}$ and the batch size of 8 for 6 epochs.
For SAM, we use the ViT-H pre-trained model. 
\begin{figure*}[!ht]
  \centering
  \includegraphics[width=\linewidth]{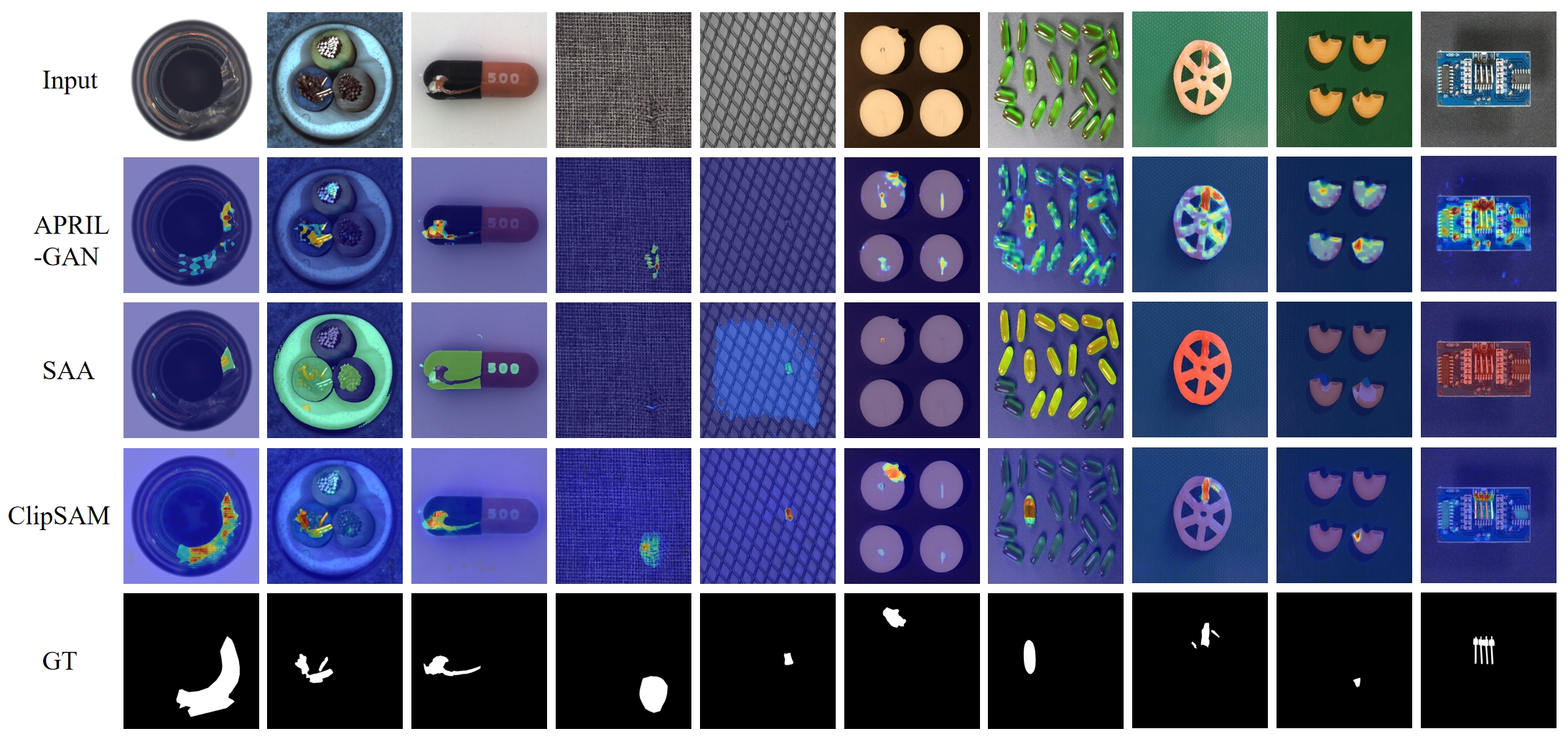}
  \vspace{-7mm}
  \caption{
  Comparison of visualization results among ClipSAM, CLIP-based, and SAM-based methods on the MVTec-AD dataset. Our ClipSAM performs much better on the location and boundary of the anomaly segmentation.
  % fine-grained segmentation.
  }
  \label{fig:Visualization}
  \vspace{-3.2mm}
\end{figure*}
\begin{figure}[!ht]
  \centering
  \includegraphics[width=0.8\linewidth]{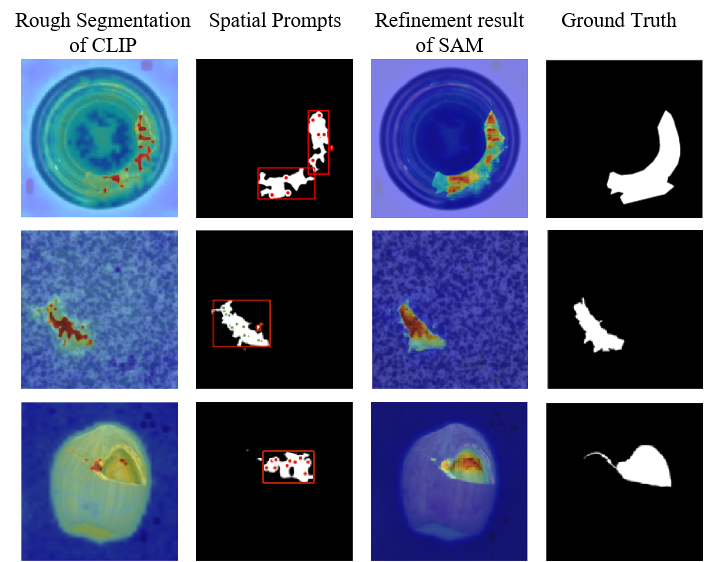}
  \vspace{-3.2mm}
  \caption{Visualization of the results of each step of our ClipSAM collaboration framework. ClipSAM first uses CLIP for rough segmentation and then uses SAM for refinement.}
  \label{fig:v3}
  \vspace{-3mm}
\end{figure}

\subsection{Experiments on MVTec-AD and VisA}
\noindent
\textbf{Comparison with state-of-the-art approaches.}
In this section, we evaluate the effectiveness of our proposed ClipSAM framework for ZSAS on the MVTec-AD and VisA datasets. 
Table \ref{tab: Quantitative Results} shows the comprehensive comparison between our proposed ClipSAM and the state-of-the-art ZSAS methods~\cite{winclip,aprilgan,saa,sdp} on different datasets and various metrics. 
% In this section, we conduct experiments utilizing our ClipSAM framework for ZSAS on the MVTec-AD dataset. % In Table \ref{tab: Quantitative Results}, we compare our approach with the state-of-the-art methods~\cite{winclip,aprilgan,saa,sdp} to demonstrate its effectiveness.
It can be concluded that our proposed ClipSAM outperforms existing state-of-the-art methods in all four metrics. Taking the MVTec-AD dataset as an example, our proposed ClipSAM outperforms the advanced CLIP-based method SDP+ by 1.1\%, 5.9\%, 6.5\% and 2.7\% on the AUROC, $F_1$-max, AP and PRO metrics respectively. Compared to the SAM-based approach, our method exhibits superior performance benefits, i.e., improvements of 19.1\%, 10.0\%, 17.1\%, and 45.5\% for the metrics. On the VisA dataset, our proposed method similarly shows an overall performance enhancement, demonstrating the effectiveness and generalization of our ClipSAM.

\noindent
\textbf{Qualitative comparisons.}
We provide some visualization of ZSAS results in Figure~\ref{fig:Visualization} to further demonstrate the effectiveness of the proposed method. For comparison, we also show the segmentation visualization of APRIL-GAN (CLIP-based method) and SAA (SAM-based method). 
It can be observed that APRIL-GAN can roughly locate the anomalies but fails to provide excellent segmentation results.
% {For example, }
In contrast, SAA can perform the segmentation well but cannot cover the anomalous region accurately. 
% {For example, }
Compared with these methods, our proposed ClipSAM provides accurate localization as well as good segmentation results. More visualization results are provided in \textbf{Appendix D}. To better understand the role of each module in the ClipSAM framework, we also visualize the rough segmentation results of the CLIP phase and the processed prompts fed into the SAM phase in Figure~\ref{fig:v3}. It shows that CLIP performs a rough segmentation of abnormal parts and generates corresponding prompts based on their locations to complete SAM's further refinement of the results.
Please refer to \textbf{Appendix E} for more details. % 这里不用进一步写出来吗？

% We show some of the visualization results in Figure~\ref{fig:Visualization}. 
% The compared approaches include APRIL-GAN, SAA, and our ClipSAM. Since the codes of WinCLIP and SDP are not open source, we choose to compare APRIL-GAN, which is another representative work of CLIP-based methods. 
% We can observe that APRIL-GAN can roughly locate the abnormal location, but performs poorly on segmentation results. 
% In addition, SAA, a SAM-based method, although its segmentation performance is well, the segmentation position is not accurate. Compared with them, ClipSAM performs well in both localization and segmentation. Please refer to Appendix D for more visualization results. 
% In addition, Figure~\ref{fig:v3} showcases the two different stages of using CLIP and SAM in our framework and qualitatively explain the process from getting rough segmentation by CLIP to refinement by SAM. Please refer to Appendix E for more details.弱化三四段对匹配，为了协同
\begin{table}[!t]
\small
% \Huge
\centering
\resizebox{\linewidth}{!}{
\begin{tabular}{cl|cccc}
\toprule
\multicolumn{2}{c|}{Components of ClipSAM} & AUROC & $F_{1}$-max & AP & PRO \\
\midrule
\multirow{2}*{UMCI} & only w/strip & 90.8 & 44.3 & 34.9 & 79.6 \\
                    & only w/scale & 90.9 & 44.4 & 42.7 & 81.9 \\
\midrule
\multirow{2}*{Module} & w/o UMCI & 60.4 & 19.7 & 25.0 & 34.7 \\
                    & w/o MMR & 91.8 & 46.7 & 44.7 & 84.8 \\
\midrule         
\multicolumn{2}{c|}{\textbf{ClipSAM(Ours)}} & \textbf{92.3} & \textbf{47.8} & \textbf{45.9} & \textbf{88.3} \\
\bottomrule
\end{tabular}
}
\vspace{-2mm}
\caption{Ablation study of different components in our ClipSAM framework on MVTec-AD dataset. Bold indicates the best results.}
\vspace{-2mm}
\label{tab: Quantitative Ablation Results for Components}
\end{table}

\begin{table}[!t]
\small
% \Huge
\centering
\resizebox{\linewidth}{!}{
\begin{tabular}{cc|cccc}
\toprule
\multicolumn{2}{c|}{Hyperparameters} & AUROC & $F_{1}$-max & AP & PRO \\
\midrule
\multirow{3}*{\makecell{Hidden dim \\ ($c_{h}$)}} & 194 & 91.8 & 45.6 & 43.8 & 83.1 \\
                       & 256 & 91.7 & 46.4 & 44.6 & 83.6 \\
                       & \textbf{384} & \textbf{92.3} & \textbf{47.8} & \textbf{45.9} & \textbf{88.3} \\        
\midrule
\multirow{3}*{\makecell{Kernel size \\ ($s_{1}$ \& $s_{2}$)}}& 2 \& 4 & 91.9 & 45.7 & 45.3 & 85.1 \\
                    & \textbf{3 \& 9} & \textbf{92.3} & \textbf{47.8} & \textbf{45.9} & \textbf{88.3} \\
                    & 6 \& 10 & 91.8 & 46.8 & 44.7 & 84.9 \\
\midrule
\multirow{3}*{\makecell{Threshold \\ ($thr$)}} & 0.45 & 91.5 & 43.3 & 44.9 & 82.5 \\
                     & \textbf{0.47} & \textbf{92.3} & \textbf{47.8} & \textbf{45.9} & \textbf{88.3} \\
                    & 0.50 & 91.7 & 44.6 & 43.4 & 83.2 \\
\bottomrule
\end{tabular}
}
\vspace{-2mm}
\caption{Ablation study of different hyperparameters used in our ClipSAM framework on MVTec-AD dataset. Bold indicates the best results in the UMCI module.  $c_{h}$ represents the hidden dimension of the convolutional layer. $s_{i}$ denotes the kernel size of the average pooling layer used in the scale path. $thr$ means the threshold for binarization. }
% $c_{h}$ represents the hidden dimension of the convolutional layer used in the UMCI module. $s_{i}$ denotes the kernel size of the average pooling layer used in the scale path. $thr$ means the threshold for binarization.}
\vspace{-2mm}
\label{tab: Quantitative Ablation Results for Hyperparameters}
\end{table}

\begin{figure}[t]
  \centering
  \includegraphics[width=0.88\linewidth]{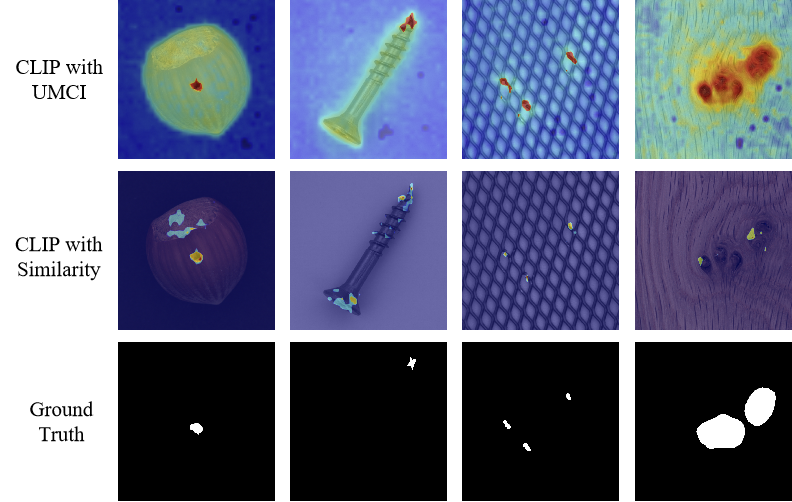}
  \vspace{-3.2mm}
  \caption{Visualization of anomaly localization and rough segmentation by CLIP with the UMCI module and CLIP with similarity calculation.}
  \label{fig:v5}
  \vspace{-3mm}
\end{figure}

\subsection{Ablation Studies}
In this section, we conduct several ablation studies on the MVTec-AD dataset to further explore the effect of different components and the experiment settings on the results in the proposed ClipSAM framework. 
% Additionally, we perform ablation studies to analyze the impact of different components and hyperparameters of our ClipSAM framework, which are displayed in Table~\ref{tab: Quantitative Ablation Results for Components} and Table~\ref{tab: Quantitative Ablation Results for Hyperparameters}, respectively.

\noindent
\textbf{Effect of components.}
Table~\ref{tab: Quantitative Ablation Results for Components} shows the results of the ablation study of different components in ClipSAM. Specifically, we first explore the impact of preserving only the strip path or the scale path in the UMCI module. Subsequently, the performance of removing either the UMCI or MMR module from the framework is tested. It can be found that removing a path in the UMCI module can lead to performance degradation, and removing the scale path has a greater impact, reflecting the necessity of combining two paths in the UMCI module. 
At the module level, removing the MMR module will slightly lower performance. Comparing this result with SDP+, we can surprisingly find that the rough segmentation results of the CLIP phase yield even better performance than CLIP-based methods. Figure~\ref{fig:v5} shows the visualization comparison between the rough segmentation and APRIL-GAN (Since SDP is not open source). The first two columns of Figure~\ref{fig:v5} indicates that UMCI can locate the anomalies more accurately, and the last two columns shows that UMCI provides better segmentation. 
In comparison with the MMR module, removing the UMCI module means regarding the similarity-based segmentation as the rough segmentation result. However, as shown in Figure~\ref{fig:Visualization}, the similarity-based segmentation cannot provide text-aligned patch tokens and accurate local spatial prompts for the subsequent MMR module. This results in a performance collapse, which demonstrates the important role of the UMCI module in the ClipSAM framework.

\noindent
\textbf{Effect of hyperparameters.} 
We explore the effects of various hyperparameters on our ClipSAM framework and record the results in Table~\ref{tab: Quantitative Ablation Results for Hyperparameters}. An analysis of the roles of each hyperparameter in the ClipSAM framework is provided based on the results.
Hidden Dimension (${c_{h}}$) determines the output feature dimension of the convolution layer. Larger ${c_{h}}$ values contribute to the effective interpretation of visual features by the model.
Kernel size (${s_{i}}$) affects the size of multi-scale visual features, which should be moderate to provide easy-to-understand visual context.
Threshold Value ($thr$) primarily impacts the initial binarized segmentation of the SAM phase. The setting of $thr$ should also be moderate: a small value may cause the non-anomalous regions to be misclassified and thus unable to generate accurate masks for SAM; a large value may cause some anomalies to be ignored and not detected.

% We illustrate the impact of different parameters on model performance under different hyperparameter settings. As illustrated in Table~\ref{tab: Quantitative Ablation Results for Hyperparameters}, the model reaches its best performance when ${c_{h}}$ is set to 384, $s_{1}$ is set to 3, $s_{2}$ is set to 9 and $thr$ is set to 0.47. Among them, ${c_{h}}$ affects the output's feature dimension of the convolutional layer and will make it difficult to understand the visual features with a small ${c_{h}}$; $s_{i}$ determines how the visual input will be transformed; $thr$ affects the quality of the binarized rough segmentation result and thereby influencing the generation of spatial prompts. If $thr$ is too small, more unexpected parts will be judged as anomalies, which will affect SAM's ability to provide accurate masks.

\section{Conclusion}
We propose the CLIP and SAM Collaboration (ClipSAM) framework to solve zero-shot anomaly segmentation for the first time. To cascade the two foundation models effectively, we introduce two modules. One is the UMCI module, which explicitly reasons anomaly locations and achieves rough segmentation. 
% by fusing visual and text features at multiple levels. 
The other is the MMR module, which refines the rough segmentation results by employing the SAM with precise spatial prompts.
% generates prompts via the clip-extracted location information to constrain the outputs of SAM and fuses them with the rough segmentation result. 
Sufficient experiments showcase that ClipSAM provides a new direction for improving ZSAS by leveraging the characteristics of different foundation models. 
In future work, we will further investigate how to integrate knowledge from different models to enhance the performance of zero-shot anomaly segmentation.
% to deal with the problem of missing samples.
%我们提出了CLIP and SAM Collaboration（ClipSAM）框架来解决零样本异常分割。为了有效级联两个模型，我们引入了两个模块。一个是UMCI模块，他通过让视觉和文本嵌入在多个级别上融合，显示地推理异常位置并实现粗分割。另一个是MMR模块，它根据位置信息生成提示来约束SAM的输出并将其与粗分割结果融合。大量的实验表明ClipSAM为ZSAS提供了新的方向，通过利用不同模型的优势来处理样本缺失的问题。

%我们提出了多模态对齐引导的动态令牌修剪（MADTP）框架来解决 VLT 的繁重计算成本。我们的 MADTP 集成了 MAG 模块，该模块跨模式调整特征并指导令牌修剪过程，以消除两种模式中不太重要的令牌。此外，还引入了DTP模块来根据输入实例的复杂性动态调整令牌压缩比。通过大量的实验，我们表明 MADTP 是一种很有前途的加速 VLT 的方法，它可以在不牺牲性能的情况下降低计算成本。

\nocite{resnet}
\nocite{anomalygpt}
\nocite{zhang2023comprehensive}
\nocite{zhang2023survey}
\nocite{yang2024foundation}
\nocite{yang2023memseg}
\nocite{chen2023revisiting}
\nocite{liu2023component}
\nocite{khattak2023self}
\nocite{ye2022b}
% \nocite{ye2022beta}
\nocite{ye2022efficient}
\nocite{ye2022stimulative}
\nocite{rafiei2023pixel}
% \nocite{simoes2024attention}
% \nocite{sui2023salvage}
% \nocite{zhang2023weakly}
% \nocite{bae2022image}

% \renewcommand\thesection{\Alph{section}}
% \setcounter{page}{1}
% \maketitlesupplementary
% \renewcommand{\thefootnote}{\daggerfootnote{footnote}}

% \section{Appendix}
\appendix
\renewcommand\thesection{\Alph{section}}
\section{Text design}
\label{sec:text}
\begin{figure*}[!htbp]
  \centering
  \includegraphics[width=0.88\linewidth]{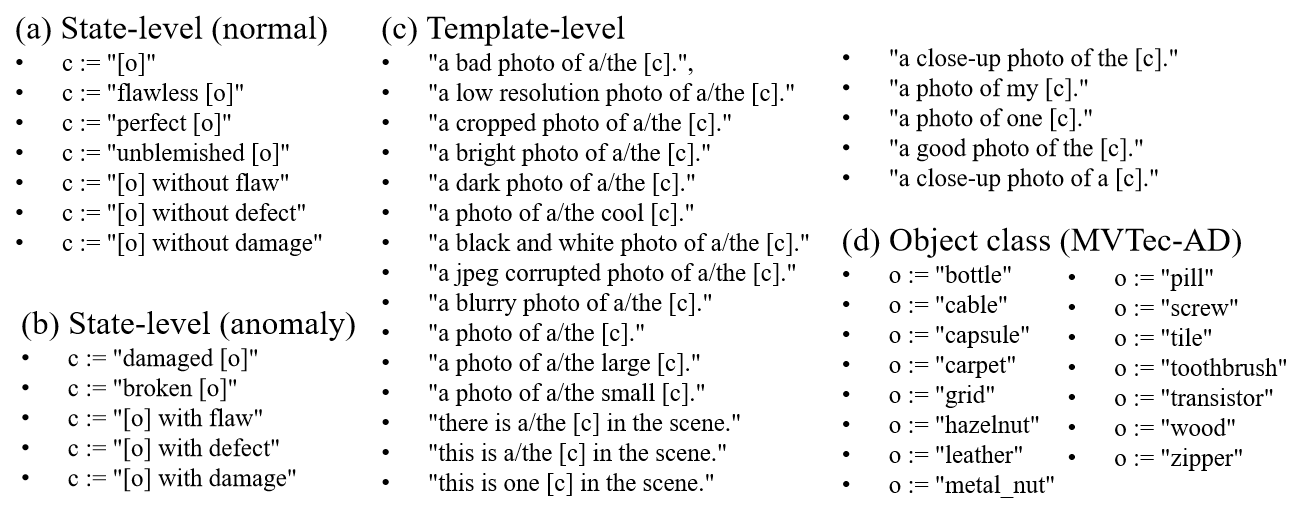}
  \vspace{-3mm}
  \caption{Lists of multi-level prompts considered in this paper to construct compositional prompt ensemble. The integrated descriptive statements are primarily categorized into phrases that describe both normal and abnormal states and templates. The category names are illustrated using the MVTec-AD dataset as example.
  }
  \label{fig:text}
  \vspace{-3mm}
\end{figure*}
As shown in Figure~\ref{fig:text}, the prompt design in ClipSAM follows the same approach in WinCLIP~\cite{winclip}. Specifically, for a given category, such as 'bottle' in the MVTec-AD dataset, phrases describing the normal state, like "perfect bottle", are combined using the category name. 
Subsequently, these phrases describing the normal state are integrated separately with the prompt templates.
This process can yield multiple descriptive statements about a normal bottle, such as "a photo of a perfect bottle." Assuming we have $m$ prompt templates and $n$ phrases describing normal states, we can generate a total of ${m\times n}$ sentences to describe a normal 'bottle'. 
To obtain text features corresponding to each sentence, we utilize the text encoder of CLIP and then compute the average of all text features that describe normality.
This yields $L_{normal}\in \mathbb{R}^{c_{t}\times 1}$. Here, $c_{t}$ represents the feature dimension, and $1$ denotes the text category, namely normal. Similarly, we can calculate the averaged feature $L_{anomaly}\in \mathbb{R}^{c_{t}\times 1}$ for sentences describing anomalies.
Finally, we concatenate $L_{normal}$ and $L_{anomaly}$ on the category dimension to obtain $L\in \mathbb{R}^{c_{t}\times 2}$, where $2$ represents the two categories, normal and abnormal.

During the training and testing processes, it is assumed that the object categories are known, while the specific anomaly categories are unknown. Consequently, for a batch of data, text features corresponding to their respective categories can be generated based on the known object categories. In situations where the categories are unknown, the placeholder 'object' can be used to substitute for specific object categories, which has been proven effective in experiments~\cite{Anomalyclip}.

\section{Strip path and Scale path}
\label{sec:ss}

The UMCI module consists of two parallel paths: the $Strip$ $Path$ and the $Scale$ $Path$. The $Strip$ $Path$ employs interactions between language features and row- and column-level visual features to calculate visually salient pixels with the strongest language perception in different directions. The $Scale$ $Path$ utilizes interactions between language features and globally visual features at different scales to comprehend what is considered anomaly.

\begin{algorithm}[!tb]
    \caption{Strip Path}
    \label{alg:Strip Path}
    \textbf{Input}: Image patch token $P$, Language feature $L$;\\
    % \textbf{Parameter}: Optional list of parameters\\
    \textbf{Output}: strip path output $M_{row,col}$
    \begin{algorithmic}[1] %[1] enables line numbers
        \STATE $\hat{P} = Linear(P)$.
        \STATE $v_{row} = conv_{1\times 3} (Avg\_Pool_{1\times W} (\hat{P}))$, \\
    $v_{col} = conv_{3\times 1} (Avg\_Pool_{H\times 1} (\hat{P}))$,\\
        \FOR{ $i$ in $(row,col)$}
        \STATE $t_{i}^{1} =  conv_{1\times 1}^{1} (L)$,\\
        $t_{i}^{2} =  conv_{1\times 1}^{2} (L)$,\\
        $t_{new}=GRU(Attention( t_{i}^{1},v_{i},v_{i}))$,\\
        $v_{i}^{att} =Attention(v_{i},t_{new},t_{i}^{2})$,\\
        $M_{i} = Norm(v_{i}^{att} + v_{i})$,\\
        \ENDFOR
        \STATE $M_{row,col} = conv_{3\times 3}(B(M_{row})+B(M_{col}))$
        \STATE \textbf{return} $M_{row,col}$
    \end{algorithmic}
\end{algorithm}

\begin{algorithm}[!tb]
    \caption{Scale Path}
    \label{alg:Scale Path}
    \textbf{Input}: Image patch token $P$, Language feature $L$;\\
    % \textbf{Parameter}: Optional list of parameters\\
    \textbf{Output}: strip path output $M_{{g}_{1},{g}_{2}}$
    \begin{algorithmic}[1] %[1] enables line numbers
        \STATE $\hat{P} = Linear(P)$.
        \STATE $v_{{g}_{1}} = conv_{3\times 3}^{{g}_{1}} (Avg\_Pool_{{s}_{1}\times {s}_{1}} (\hat{P}))$, \\
        $v_{{g}_{2}} = conv_{3\times 3}^{{g}_{2}} (Avg\_Pool_{{s}_{2}\times {s}_{2}} (\hat{P}))$,\\
        \FOR{ $i$ in $({g}_{1},{g}_{2})$}
        \STATE $t_{i}^{k} =  conv_{1\times 1}^{k} (L)$  , \\
        $t_{i}^{v} =  conv_{1\times 1}^{v} (L)$,\\
        $v_{i}^{att}=Attention(v_{i},text_{i}^{k},text_{i}^{v})$,\\
        $M_{i} = Norm(v_{i}^{att}+v_{i})$,
        \ENDFOR
        \STATE $M_{{g}_{1},{g}_{2}} = conv_{3\times 3}^{{g}_{1},{g}_{2}}(B(M_{{g}_{1}})+B(M_{{g}_{2}}))$
        \STATE \textbf{return} $M_{{g}_{1},{g}_{2}}$
    \end{algorithmic}
\end{algorithm}

\begin{figure}[!htbp]
  \centering
  \includegraphics[width=0.88\linewidth]{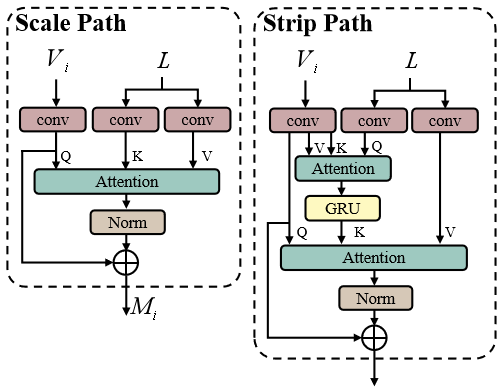}
  \vspace{-3mm}
  \caption{Diagram depicting the attention calculation process in the strip path and scale path.
  }
  \label{fig:att}
  \vspace{-3mm}
\end{figure}

\noindent
(1) \textbf{${Strip}$ ${Path}$}. 
The interaction conducted by the strip path involves the fusion of text features and row-and-column visual features. Specifically, visual features are aggregated into row-level and column-level features from horizontal and vertical directions through average pooling layers. Taking row-level features $v_{row}$ as an example, text features $L$ are processed by convolutional layers to obtain language features $t_{row}^{1}$, $t_{row}^{2}$ with dimensions matching the visual features. Subsequently, a two-stage attention mechanism is employed to perceive relevant language features at each pixel of the row-level features. 
In the attention computation process, we employ the Scaled Dot-Product Attention mechanism~\cite{transformer}: 
\begin{equation}
\begin{aligned}
    &Attention ( Q,K,V )=softmax ( \frac{QK^{T} }{\sqrt{d_{k} } } )V.
\end{aligned}
\end{equation}
As shown in Figure~\ref{fig:att}, the first attention step is designed to capture the correlated visual features corresponding to each language feature. Then we use GRU ~\cite{gru} to merge the learned visual features with the original language features, which we can obtain language features enriched with visual information. Taking this new language feature, original visual and language features as K, Q, V respectively for attention computation can effectively aggregating features. Finally, we use the residual method to add $v_{row}$ to the result and get $M_{row}\in \mathbb{R}^{H \times c_{h}}$:
\begin{equation}
\begin{aligned}
    &t_{new}=GRU(Attention( t_{row}^{1},v_{row},v_{row})),\\
    &v_{row}^{att} =Attention(v_{row},t_{new},t_{row}^{2}),\\
    &M_{row} = Norm(v_{row}^{att} + v_{row}),
\end{aligned}
\end{equation}
where ${Norm}$ denotes L2 regularization. Please refer to Algorithm~\ref{alg:Strip Path} for specific content.

\noindent
(2) \textbf{${Scale}$ ${Path}$}.
The scale path is utilized for the interaction between text and multi-scaled visual features. Initially, visual features at different scales are obtained through average pooling layers with different kernel sizes. Taking one of these scales as an example, convolutional layers are employed to process text features to match the feature dimensions. Subsequently, a dedicated attention mechanism is employed for cross-modal interaction to obtain $M_{g_{1}}\in \mathbb{R}^{h_{g_{1}} \times w_{g_{1}} \times c_{h}}$. The residual connection is also used here to add the original vision and the result. The specific formula is as follows:
\begin{equation}
\begin{aligned}
    &v_{{g}_{1}}^{att}=Attention(v_{{g}_{1}},text_{{g}_{1}}^{k},text_{{g}_{1}}^{v}),\\
    &M_{{g}_{1}} = Norm(v_{{g}_{1}}^{att}+v_{{g}_{1}}),
\end{aligned}
\end{equation}
Please refer to Algorithm~\ref{alg:Scale Path} for specific content.

It is worth noting that all convolutional layers in the UMCI module are independent of each other.

\begin{table*}[!ht]
    \small
    \centering  
  \begin{tabular}{lccccc|cccc}
    \toprule
    &  &    \multicolumn{4}{c|}{MTD}& \multicolumn{4}{c}{KSDD2}  \\ 
\multirow{-2}{*}{Base model}  & \multirow{-2}{*}{Method}  & \multicolumn{1}{c}{AUROC}   & \multicolumn{1}{c}{$F_{1}$-max}   & \multicolumn{1}{c}{AP} & \multicolumn{1}{c|}{PRO}   & \multicolumn{1}{c}{AUROC}   & \multicolumn{1}{c}{$F_{1}$-max}   & \multicolumn{1}{c}{AP} & \multicolumn{1}{c}{PRO}          \\ 
\midrule
 \multirow{1}{*}{CLIP-based Approaches} 
 & APRIL-GAN & 51.1 & 16.6 & 9.8 & 17.2 & 52.1 & 10.7 & 9.3 & 13.3 \\\midrule
 \multirow{1}{*}{SAM-based Approaches}& SAA+ & 69.4 & 37.3 & 28.8 & - & 77.5 & 61.6 & 49.6 & - \\
\midrule
 CLIP \& SAM& \textbf{ClipSAM(Ours)} & \textbf{88.0} & \textbf{55.2} & \textbf{51.9} & \textbf{71.3} & \textbf{90.7} & \textbf{67.2} & \textbf{67.9} & \textbf{88.8} \\
    \bottomrule
  \end{tabular}
 \vspace{-1mm}
  \caption{Performance comparison of different kinds of ZSAS approaches on the MTD and KSDD2 datasets. Evaluation metrics include AUROC, $F_{1}$-max, AP, and PRO. 
% under zero-shot setting. 
Bold indicates the best results.
  }
  \vspace{2mm}
\label{tab:mtd_ksdd2}
\end{table*}

\begin{figure*}[!ht]
  \centering
  \includegraphics[width=\linewidth]{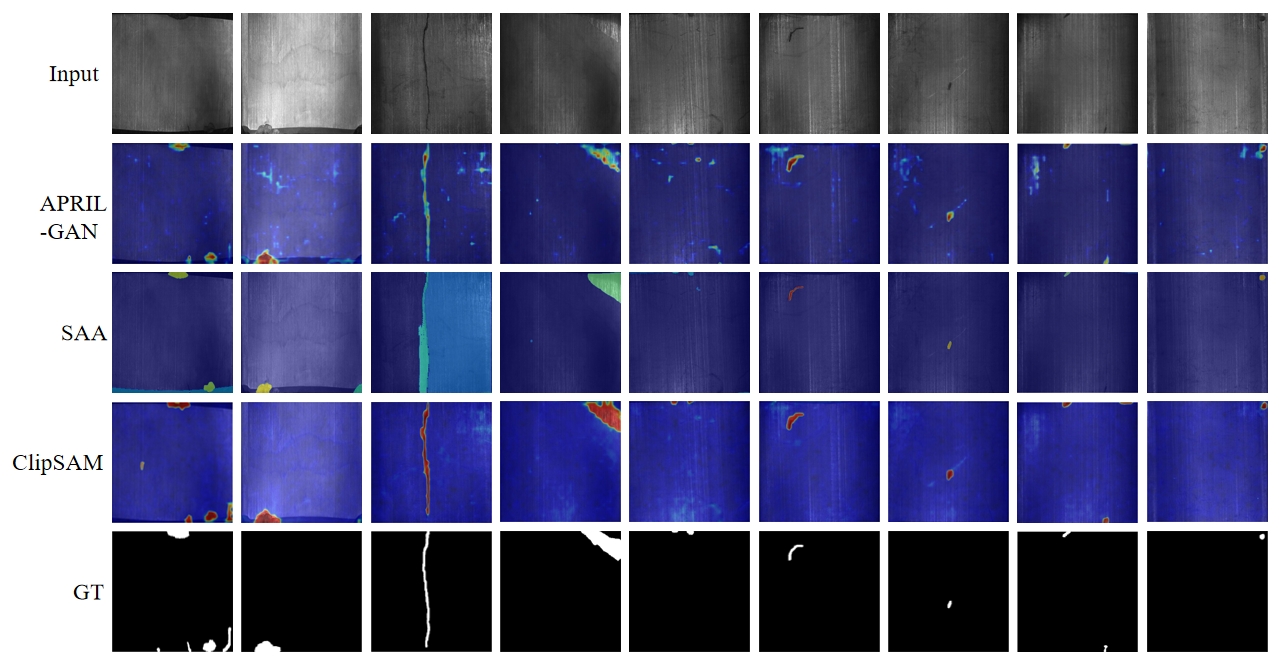}
  \vspace{-5mm}
  \caption{
  Comparison of visualization results among ClipSAM, CLIP-based, and SAM-based methods on the MTD dataset. Our ClipSAM performs much better on the location and boundary of the anomaly segmentation.
  % fine-grained segmentation.
  }
  \label{fig:mtd}
  \vspace{-3mm}
\end{figure*}

\section{Additional experiments on more datasets}
\label{sec:ad}
We validated the effectiveness of ClipSAM on two commonly used datasets for zero-shot anomaly segmentation, namely MVTec-AD~\cite{mvtec-ad} and ViSA~\cite{visa}. Additionally, we conducted relevant experiments on the MTD~\cite{mtd} and KSDD2~\cite{ksdd2} datasets to further confirm the generalizability of our approach.
\subsection{MVTec-AD}
The MVTec-AD dataset serves as an unsupervised anomaly detection dataset, comprising a total of 3466 unlabeled images and 1888 annotated images with pixel-level segmentation annotations. The image sizes are either 700×700 or 1024×1024 pixels. The training dataset consists of 3629 images, all depicting defect-free instances. The test dataset comprises 1725 images, including both defective and defect-free samples. The dataset encompasses 15 categories, comprising 5 texture categories such as carpets and leather, and 10 object categories including bottles, cables, capsules, chestnuts, and others. In total, the dataset contains 73 types of anomalies, such as scratches, dents, and missing parts. Standard evaluation metrics, including AUROC, AP, $F_{i}$-max and PRO are commonly employed for assessment.
\subsection{VisA}
The VisA dataset is also one of the commonly used datasets for zero-shot anomaly segmentation. It consists of 10,821 images, comprising 9,621 normal samples and 1,200 anomaly samples. The dataset is organized into 12 subsets, each corresponding to a different object category. Among them, four subsets represent different types of printed circuit boards (PCBs) with relatively complex structures, including transistors, capacitors, chips, and other components. Additionally, four subsets (Capsules, Candles, Macaroni1, and Macaroni2) contain multiple instances in their views. Instances in Capsules and Macaroni2 exhibit significant variations in both position and orientation. The performance of the model on this dataset can also be evaluated using AUROC, AP, $F_{i}$-max and PRO.
\subsection{MTD}
The Magnetic Tile Defect (MTD) dataset is a more specialized dataset that consists of images related to a single object category but with various defect categories. It encompasses six common defect types in magnetic tiles, such as blowhole, break, crack, and others. The dataset comprises a total of 925 images without defects and 392 images with anomalies, each with corresponding image annotations. Performance evaluation can be conducted using the AUROC, AP, $F_{i}$-max and PRO metrics in a similar manner.
\subsection{KSDD2}
The Kolektor Surface-Defect Dataset 2 (KSDD2) is relevant to industrial quality inspection and can also be used for anomaly segmentation. It comprises 356 images with obvious defects and 2979 images without any defects. The dimensions of each image in the dataset are approximately 230×630 pixels. The dataset is divided into a training set and a test set, with the training set consisting of 246 positive images and 2085 negative images, while the test set includes 110 positive images and 894 negative images. The dataset encompasses various types of defects, including scratches, small spots, surface defects, and others. The performance of the model in zero-shot anomaly segmentation on this dataset can also be assessed using the AUROC, $F_{1}$-max, AP, and PRO metrics.
\subsection{Experimental results}
\textbf{Implementation details.} Our experiments adhere to the settings defined in AnomalyCLIP~\cite{Anomalyclip}. Specifically, the model is trained on the MVTec-AD dataset and tested on the MTD and KSDD2 datasets. For the MTD dataset, all data samples are considered as the test set. In the case of the KSDD2 dataset, we directly utilize its test set.

In the experiments, the pre-trained ViT-L-14-336 model released by OpenAI, which consists of 24 Transformer layers, is utilized for CLIP encoders. We extracted the image patch tokens after each stage of the image encoder (i.e., layers 6, 12,18, and 24) for the training of our proposed UMCI module, respectively. The optimization process is conducted on a single NVIDIA 3090 GPU using AdamW optimizer with the learning rate of ${1\times 10^{-4}}$ and the batch size of 8 for 6 epochs.
For SAM, we use the ViT-H pre-trained model.

\begin{figure*}[!ht]
  \centering
  \includegraphics[width=\linewidth]{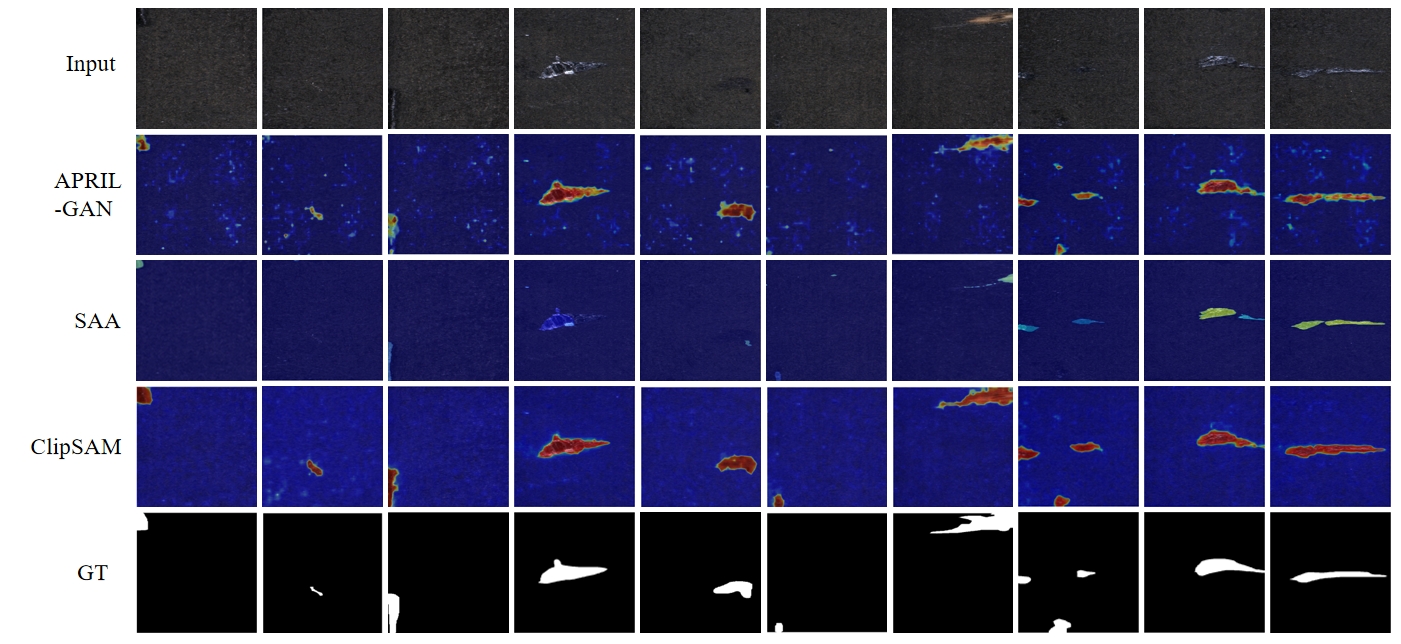}
  \vspace{-5mm}
  \caption{
  Comparison of visualization results among ClipSAM, CLIP-based, and SAM-based methods on the KSDD2 dataset. Our ClipSAM performs much better on the location and boundary of the anomaly segmentation.
  % fine-grained segmentation.
  }
  \label{fig:ksdd2}
  \vspace{-3mm}
\end{figure*}

\noindent
\textbf{Comparison with state-of-the-art approaches.} Due to the fact that CLIP-based methods such as SDP~\cite{sdp} are not open source and have not been experimentally evaluated on the MTD and KSDD2 datasets, no comparison is provided here. Therefore, we evaluate the performance of the CLIP-based method APRIL-GAN~\cite{aprilgan}, the SAM-based method SAA~\cite{saa}, and our proposed ClipSAM on the zero-shot anomaly segmentation task. As shown in Table~\ref{tab:mtd_ksdd2}, our method demonstrates significant improvements over the other two methods across different metrics. In particular, taking the MTD dataset as an example, when compared to APRIL-GAN, ClipSAM demonstrates improvements of 36.9\% in AUROC, 38.6\% in $F_{1}$-max, 42.1\% in AP, and 54.1\% in PRO. When contrasted with SAA+, ClipSAM shows enhancements of 18.6\% in AUROC, 17.9\% in $F_{1}$-max, and 23.1\% in AP. Additionally, ClipSAM also demonstrates remarkable performance advantages on the KSDD2 dataset.

\noindent
\textbf{Qualitative comparisons.} We further visualize the experimental results on the MTD dataset and KSDD2 dataset in Figure~\ref{fig:mtd} and Figure~\ref{fig:ksdd2}. The compared methods include APRIL-GAN~\cite{aprilgan} (CLIP-based method), SAA~\cite{saa} (SAM-based method), and ClipSAM. Similar conclusions can be observed from both figures, indicating that ClipSAM outperforms the other two methods in terms of localization and segmentation capabilities. Meanwhile, APRIL-GAN and SAA exhibit similar disadvantages on both datasets. Specifically, the CLIP-based method, although capable of roughly locating the position of defects guided by language, suffers from inaccurate segmentation areas. Additionally, unexpected high predictions occur outside the real masks, as shown in the third and fifth columns of Figure~\ref{fig:mtd}. This is attributed to misalignment between image patch tokens and text tokens, resulting in erroneous predictions and a decrease in performance.

Furthermore, while the SAM-based method demonstrates good segmentation performance, it generates incorrect segmentation masks due to the use of ambiguous positional information words as prompts. In particular, in the third column of Figure~\ref{fig:mtd}, SAA produces large-area erroneous masks, and in the eighth column of Figure~\ref{fig:ksdd2}, SAA misses parts of the masks. In contrast, ClipSAM performs well on the entire dataset, accurately locating anomalous positions and generating precise segmentation masks without exhibiting errors in unexpected locations, as seen in APRIL-GAN.

\begin{figure*}[!ht]
  \centering
  \includegraphics[width=\linewidth]{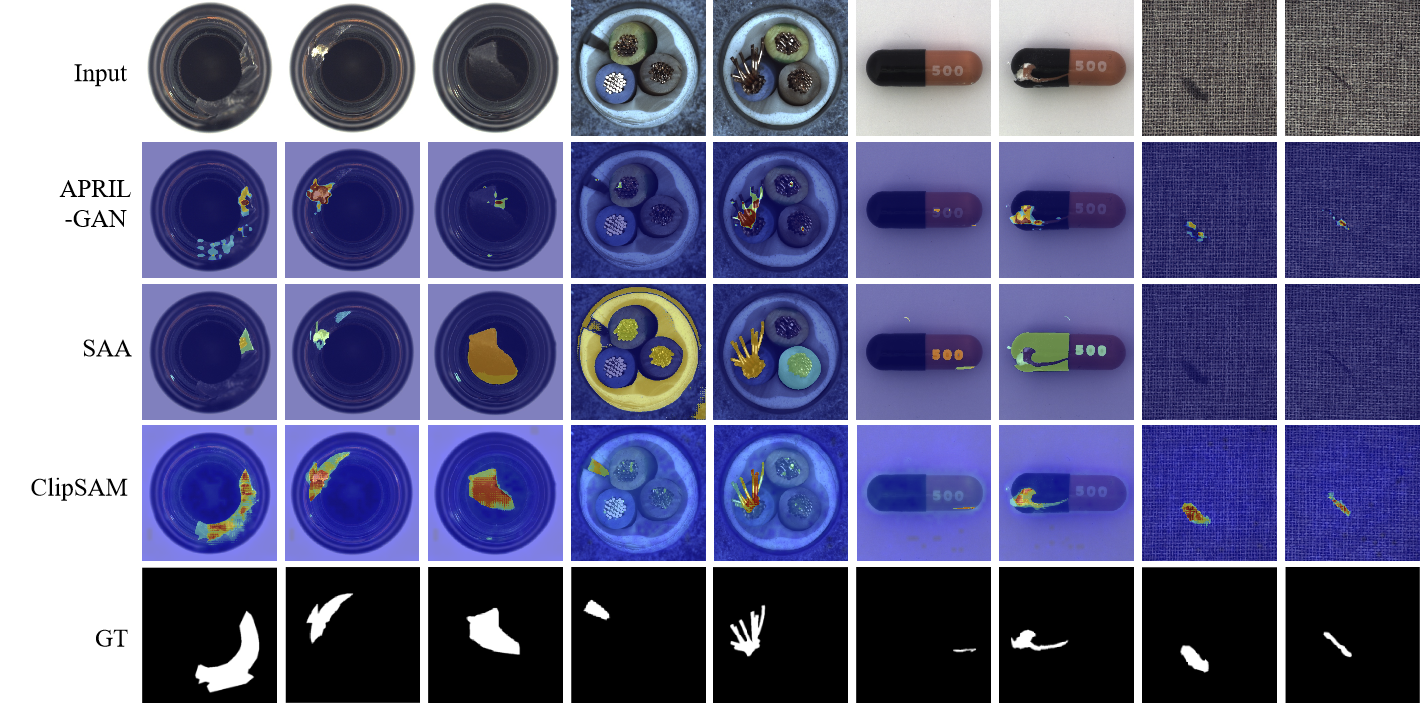}
  \vspace{-3mm}
  \caption{
  Comparison of visualization results among ClipSAM, CLIP-based, and SAM-based methods on the MVTec-AD dataset. Our ClipSAM performs much better on the location and boundary of the anomaly segmentation.
  }
  \label{fig:mvtec3}
  \vspace{-2mm}
\end{figure*}

\begin{figure*}[!ht]
  \centering
  \includegraphics[width=\linewidth]{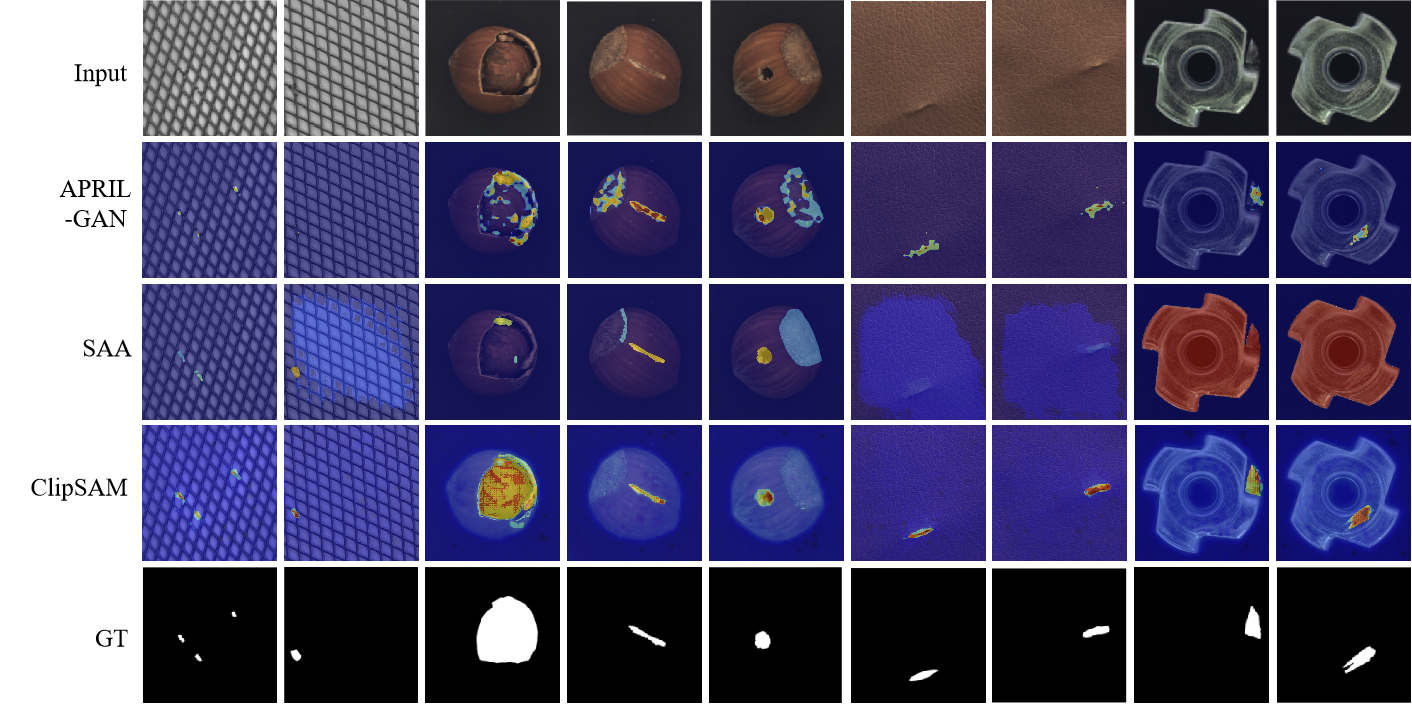}
  \vspace{-5mm}
  \caption{
 Comparison of visualization results among ClipSAM, CLIP-based, and SAM-based methods on the MVTec-AD dataset. Our ClipSAM performs much better on the location and boundary of the anomaly segmentation.
  }
  \label{fig:mvtec4}
\end{figure*}

\begin{figure*}[!ht]
  \centering
  \includegraphics[width=\linewidth]{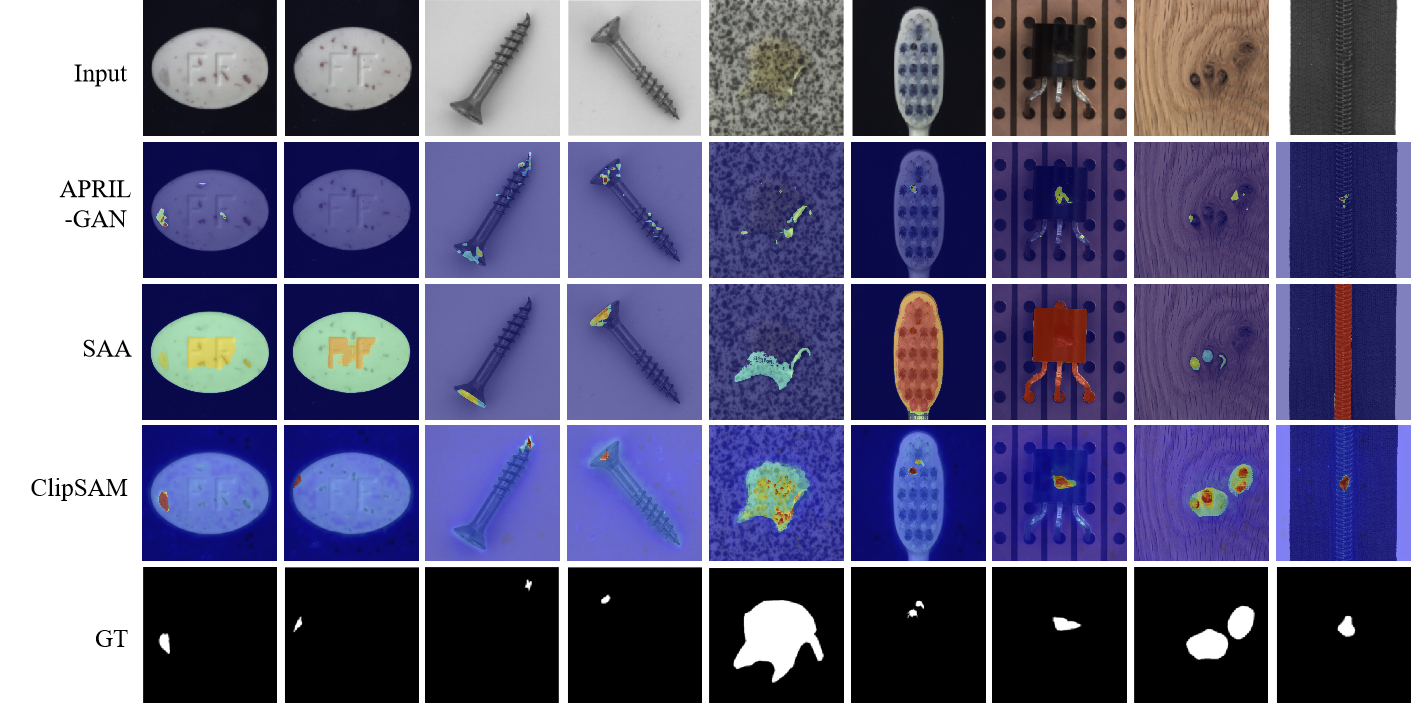}
  \vspace{-5mm}
  \caption{
  Comparison of visualization results among ClipSAM, CLIP-based, and SAM-based methods on the MVTec-AD dataset. Our ClipSAM performs much better on the location and boundary of the anomaly segmentation.
  }
  \label{fig:mvtec5}
  \vspace{-1mm}
\end{figure*}

\begin{figure*}[!ht]
  \centering
  \includegraphics[width=\linewidth]{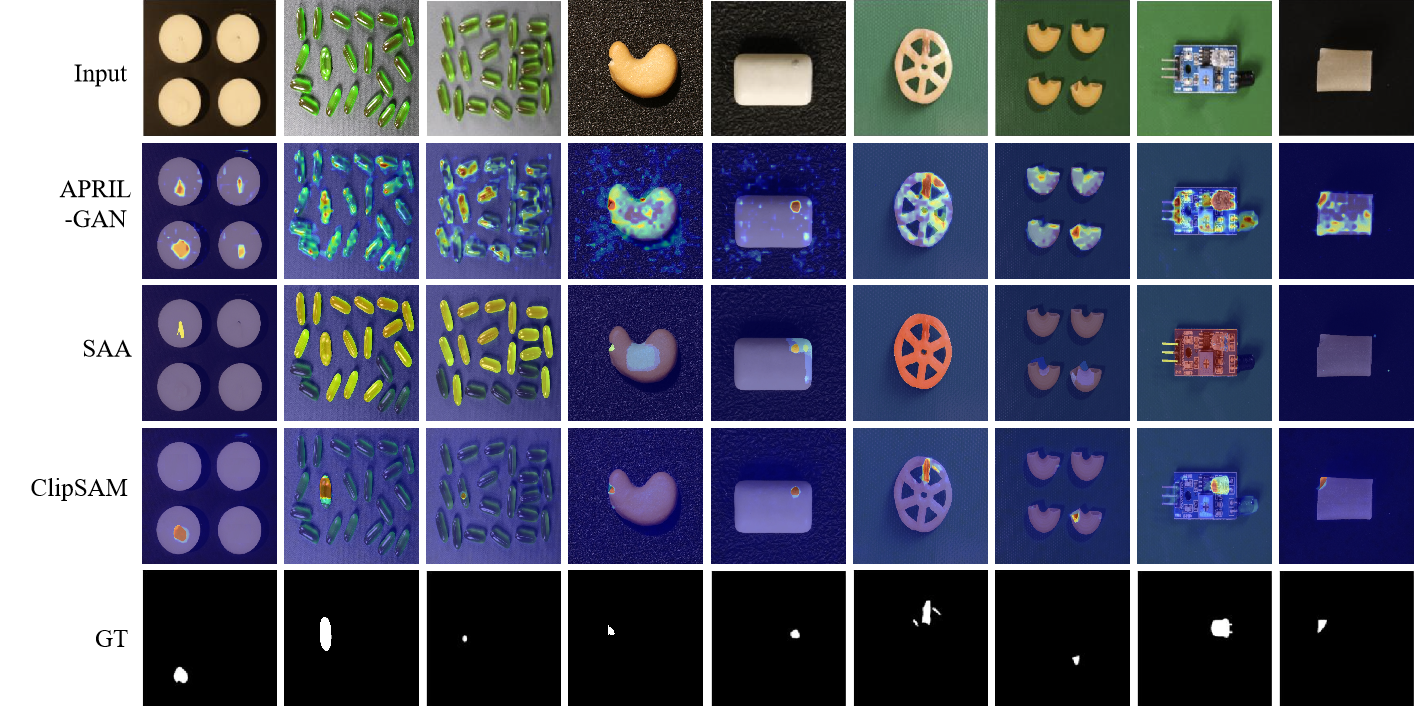}
  \vspace{-5mm}
  \caption{
  Comparison of visualization results among ClipSAM, CLIP-based, and SAM-based methods on the VisA dataset. Our ClipSAM performs much better on the location and boundary of the anomaly segmentation.
  }
  \label{fig:visa2}
  \vspace{-3mm}
\end{figure*}

\section{Additional visualization}
\label{sec:av}
In this section, we visualize the anomaly segmentation results of the proposed ClipSAM framework on the MVTec-AD dataset and the VisA dataset.
Figure~\ref{fig:mvtec3}, ~\ref{fig:mvtec4}, ~\ref{fig:mvtec5}, ~\ref{fig:visa2} illustrate the visual comparisons between APRIL-GAN~\cite{aprilgan} (CLIP-based method), SAA~\cite{saa} (SAM-based method), and our ClipSAM method. Clearly, our ClipSAM demonstrates stronger understanding of anomalous regions, benefiting from the designed UMCI and MMR modules. The process of rough segmentation by CLIP followed by refinement using SAM successfully mitigates misdetections and omissions of certain anomalies.

Specifically, CLIP-based methods tend to incorrectly classify regions outside actual anomaly areas as anomalies. Even when identifying anomaly locations, their segmentation results often deviate significantly from the ground truth labels. On the other hand, SAM-based methods heavily rely on post-processing steps for masks. While the initial candidate masks may contain correct masks, complex filtering can introduce substantial biases. Additionally, due to the vague semantic descriptions guiding the model's attention, SAM might focus on parts outside the anomaly regions and segment them entirely, as shown in the fourth column of Figure~\ref{fig:mvtec3}. In reality, anomalies usually constitute a small portion of the entire object, and such errors can significantly impact the results.

Observing the figures, APRIL-GAN tends to identify anomaly locations, though less accurately. SAM provides accurate segmentation of components but lacks precise constraints, leading to significant deviations in results. In contrast, ClipSAM's two-stage strategy effectively combines the strengths of CLIP and SAM, resulting in better performance in zero-shot anomaly segmentation.

\section{Two-stage visualization}
\label{sec:tv}

\begin{figure*}[!ht]
  \centering
  \includegraphics[width=\linewidth]{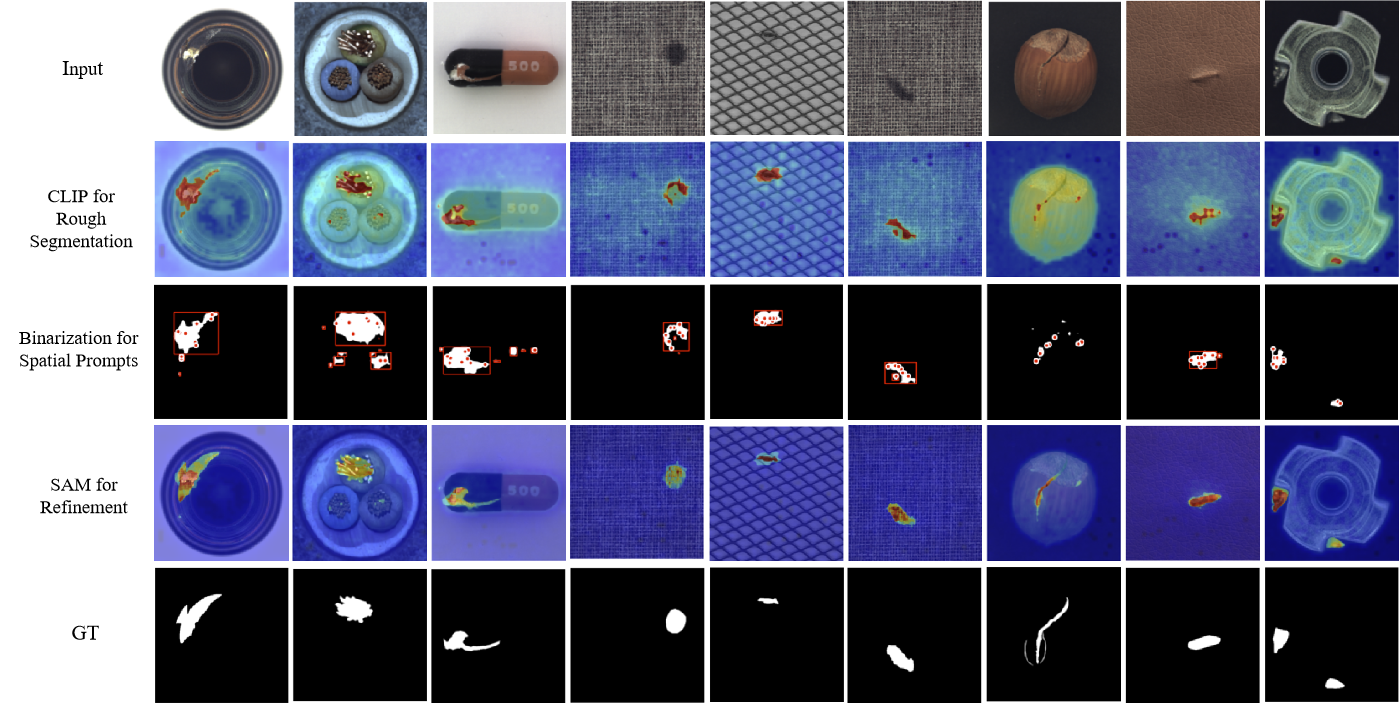}
  \vspace{-5mm}
  \caption{
  Visualization of different stages (the process of localization and rough segmentation by CLIP followed by result refinement through SAM) under the ClipSAM framework on the MVTec-AD dataset.
  }
  \label{fig:mvtec1}
  \vspace{-2mm}
\end{figure*}

\begin{figure*}[!ht]
  \centering
  \includegraphics[width=\linewidth]{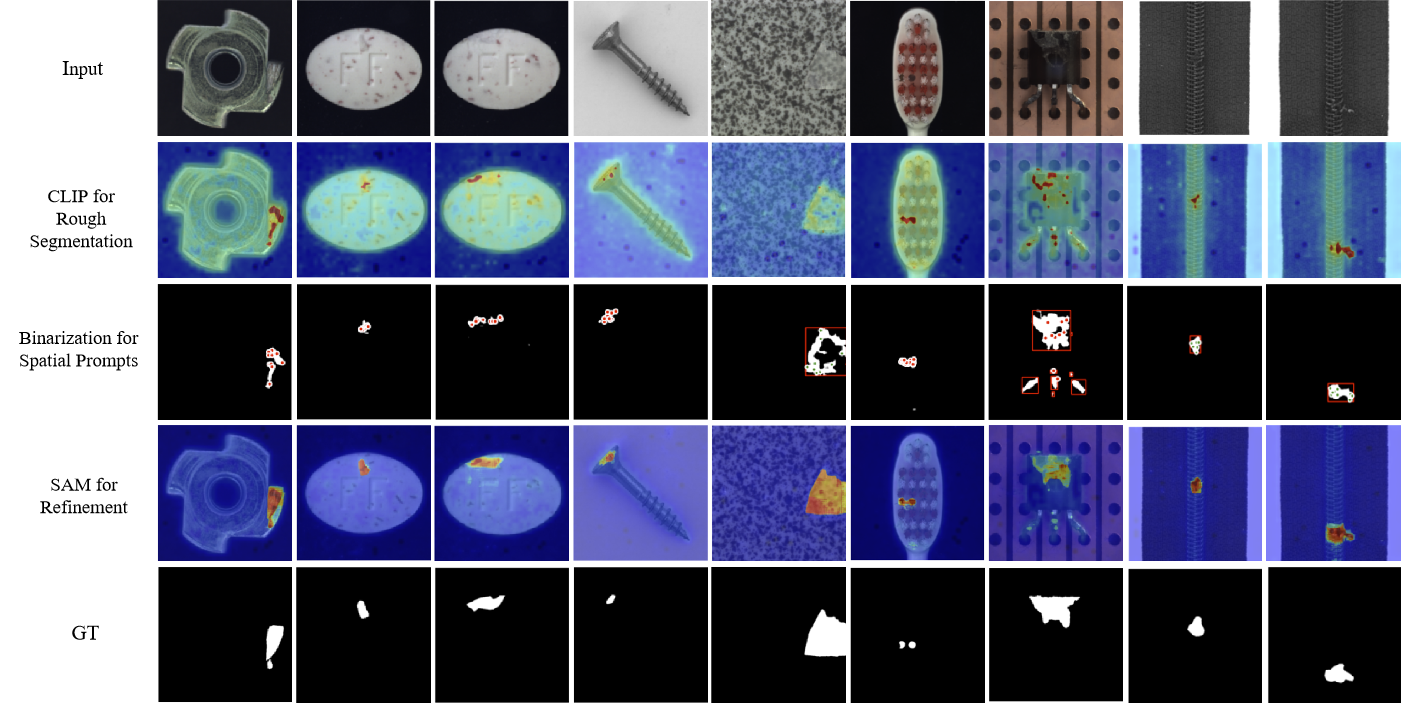}
  \vspace{-5mm}
  \caption{
  Visualization of different stages (the process of localization and rough segmentation by CLIP followed by result refinement through SAM) under the ClipSAM framework on the MVTec-AD dataset.
  }
  \label{fig:mvtec2}
  \vspace{-1mm}
\end{figure*}

\begin{figure*}[!ht]
  \centering
  \includegraphics[width=\linewidth]{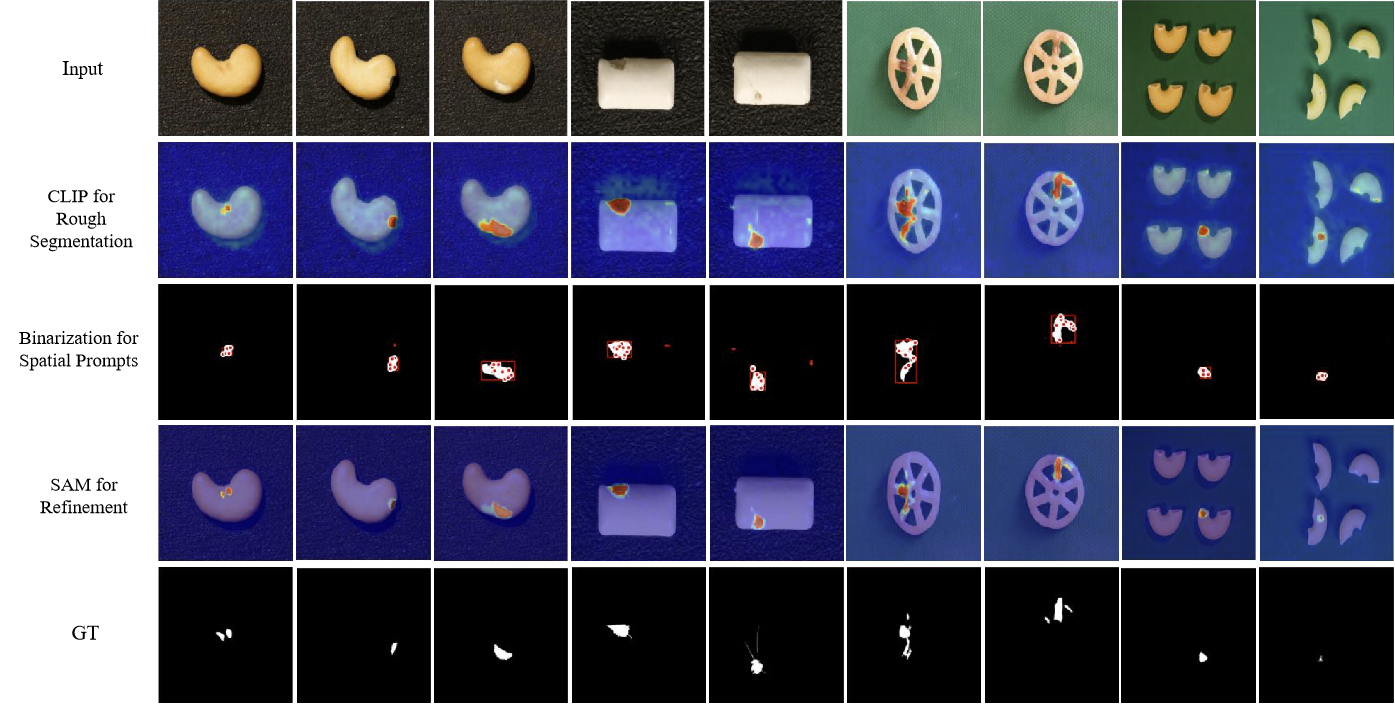}
  \vspace{-4mm}
  \caption{
  Visualization of different stages (the process of localization and rough segmentation by CLIP followed by result refinement through SAM) under the ClipSAM framework on the VisA dataset.
  }
  \label{fig:visa1}
  \vspace{2mm}
\end{figure*}

\begin{table*}[!htbp]
\small
% \Huge
\centering
\resizebox{0.9\linewidth}{!}{
\begin{tabular}{l|l|cccc|cccc}
\toprule
&  &    \multicolumn{4}{c|}{MVTec-AD}& \multicolumn{4}{c}{VisA}  \\ 
\multirow{-2}{*}{Base model}  & \multirow{-2}{*}{Method}  & \multicolumn{1}{c}{AUROC}   & \multicolumn{1}{c}{$F_{1}$-max}   & \multicolumn{1}{c}{AP} & \multicolumn{1}{c|}{PRO}   & \multicolumn{1}{c}{AUROC}   & \multicolumn{1}{c}{$F_{1}$-max}   & \multicolumn{1}{c}{AP} & \multicolumn{1}{c}{PRO}          \\ 

% \multirow{2}*{Base model} & \multirow{2}*{Method} & \multicolumn{4}{c|}{MVTec-AD}  & \multicolumn{4}{c}{VisA} \\
% \cmidrule[0.6pt](r){3-6} \cmidrule[0.6pt](r){7-10}
% & & AUROC & $F_{1}$-max & AP & PRO & AUROC & $F_{1}$-max & AP & PRO \\
\midrule
 \multirow{6}{*}{\makecell{CLIP-based \\ Approaches}} & WinCLIP & 85.1 & 31.7 & - & 64.6 & 79.6 & 14.8 & - & 56.8 \\
 & APRIL-GAN & 87.6 & 43.3 & 40.8 & 44.0 & 94.2 & 32.3 & 25.7 & 86.8 \\
 & AnoVL & 90.6 & 36.5 & - & 77.8 & 91.4 & 17.4 & - & 75.0 \\
 & AnomalyCLIP & 91.1 & - & - & 81.4 & 95.5 & - & - & 87.0 \\
 & SDP & 88.7 & 35.3 & 28.5 & 79.1 & 84.1 & 16.0 & 9.6 & 63.4 \\
 & SDP+ & 91.2 & 41.9 & 39.4 & 85.6 & 94.8 & 26.5 & 20.3 & 85.3 \\ \midrule
 \multirow{2}{*}{\makecell{SAM-based \\ Approaches}}& SAA & 67.7 & 23.8 & 15.2 & 31.9 & 83.7 & 12.8 & 5.5 & 41.9 \\
 & SAA+ & 73.2 & 37.8 & 28.8 & 42.8 & 74.0 & 27.1 & 22.4 & 36.8 \\
\midrule
 CLIP \& SAM& \textbf{ClipSAM(Ours)} & \textbf{92.3} & \textbf{47.8} & \textbf{45.9} & \textbf{88.3} & \textbf{95.6} & \textbf{33.1} & \textbf{26.0} & \textbf{87.5} \\
\bottomrule
\end{tabular}
}
\vspace{1mm}
\caption{Performance comparison of different kinds of ZSAS approaches on the MVTec-AD and VisA datasets. Evaluation metrics include AUROC, $F_{1}$-max, AP, and PRO. 
% under zero-shot setting. 
Bold indicates the best results.}
\vspace{-3mm}
\label{tab: moreQuantitative Results}
\end{table*}

The ClipSAM framework consists of two stages. In the first stage, CLIP is employed for localization and rough segmentation, while the second stage utilizes SAM for refining the results. During the process, we binarize the rough segmentation output of CLIP to generate spatial prompts as constraints in connected regions, namely points and boxes. We visualize each of these steps separately to qualitatively observe the output of each stage of the model. 

As illustrated in Figure~\ref{fig:mvtec1} and ~\ref{fig:mvtec2}, with the assistance of the Unified Multi-scale Cross-modal Interaction (UMCI) module, CLIP achieves rough segmentation of abnormal regions. However, due to CLIP learning multi-modal semantics tailored for classification tasks, it has limitations in fine-grained segmentation tasks. This is typically manifested as CLIP correctly predicting only parts of a given real anomalous region. With the aid of SAM, it is possible to further refine the abnormal regions predicted by CLIP, obtaining abnormal masks that are closer to ground truth values. It is noteworthy that the binarization of results is based on a certain threshold, which usually does not result in a one-to-one correspondence between the red regions in the rough segmentation output and the connected regions in the binarized mask. 

Additionally, as shown in Figure~\ref{fig:visa1}, due to the typically small size of anomaly categories in the ViSA dataset, the designed UMCI module has already assisted CLIP in achieving accurate anomaly segmentation. However, these results often exceed the ground truth values, as seen in the second column of the figure. SAM aids in further focusing on these small anomalous areas. After merging the masks generated by SAM with the rough segmentation results, a certain value suppression is applied to regions beyond the ground truth, which is advantageous when computing metrics.

\section{Comparing with more methods.}
\label{sec:cm}
We additionally compared the model's performance on zero-shot anomaly segmentation with two other works, AnoVL~\cite{AnoVL} and AnomalyCLIP~\cite{Anomalyclip}. AnoVL enhances the prompt templates with domain-specific textual designs such as "industrial" and "manufacturing." AnomalyCLIP introduces the concept of prompt learning to the text encoder part of CLIP. As shown in the Table~\ref{tab: moreQuantitative Results}, our approach still achieves optimal performance on the zero-shot anomaly segmentation task. However, it is noteworthy that the design for text generality is an interesting approach for addressing zero-shot anomaly segmentation tasks. ClipSAM adopts the text strategy from WinCLIP~\cite{winclip} without further modifications, presenting a potential challenge for future exploration.
% 所有向主轨道提交的论文必须最多七页，加上最多两页的参考文献/致谢/贡献声明/道德声明。作者信息匿名（不需要致谢和贡献声明）。作者信息放paper id。摘要不超过200字。新段落要缩进，主标题之后除外。有时候可能存在一些文献是确实被引用但在文本中没有直接提到的情况。这时，可以使用\nocite{}命令将这些未明确引用的文献加入到生成的参考文献列表中。在插图和标题覆盖的区域周围留出 1/4 英寸的边距。 使用 9 号字体作为插图中的标题、标签和其他文本。 说明文字应始终出现在插图下方。表格三线表，数字列右对齐，更容易比较数字。 确保还右对齐相应的标题，并对所有数字使用相同的精度。我们避免行与行之间（在本例中无需重复场景名称）以及内容中（单位可以显示在列标题中）中不必要的重复。长公式拆成两行，严禁改字体大小

%% The file named.bst is a bibliography style file for BibTeX 0.99c
\bibliographystyle{named}
\bibliography{ijcai24}

\end{document}